\documentclass[runningheads]{llncs}

\usepackage{eccv}

\usepackage{eccvabbrv}

\usepackage{graphicx}
\usepackage{booktabs}
\usepackage{wrapfig}

\usepackage[accsupp]{axessibility}  
\usepackage{adjustbox}
\usepackage{physics}
\usepackage{multirow}

\newcommand\T{\rule{0pt}{2.5ex}}       
\newcommand\B{\rule[-1.2ex]{0pt}{0pt}} 

\usepackage{amsmath,amssymb}
\usepackage[fleqn]{nccmath}
\usepackage{makecell}
\usepackage{bm}

\newcommand{\cL}{\mathcal{L}}
\newcommand{\tx}{\tilde{x}}
\newcommand\blfootnote[1]{%
  \begingroup
  \renewcommand\thefootnote{}\footnote{#1}%
  \addtocounter{footnote}{-1}%
  \endgroup
}

\usepackage{hyperref}

\usepackage{orcidlink}

\begin{document}

\title{On the Vulnerability of Skip Connections to Model Inversion Attacks} 


\author{Koh Jun Hao$^*$ \and
Sy-Tuyen Ho$^*$ \and
Ngoc-Bao Nguyen \and
Ngai-Man Cheung}

\authorrunning{Koh et al.}

\institute{Singapore University of Technology and Design (SUTD) 
\email{\{junhao\_koh,ngaiman\_cheung\}@sutd.edu.sg}}

\maketitle

\blfootnote{$^*$ Co-first authors}

\begin{abstract}
Skip connections are fundamental architecture designs for modern deep neural networks (DNNs) 
such as CNNs and ViTs. 
While they help improve model performance significantly, we identify a vulnerability
associated with skip connections to Model Inversion (MI) attacks, a type of privacy attack that aims to reconstruct private training data through abusive exploitation of a model. In this paper, as a pioneer work to understand how DNN architectures affect MI, we study the impact of skip connections on MI. \textbf{We make the following discoveries:} 1) Skip connections reinforce MI attacks and compromise data privacy. 
2) Skip connections in the {\em last stage} are the most critical to attack. 
3) RepVGG, an approach to remove skip connections in the inference-time architectures, could not mitigate the vulnerability to MI attacks. 
4) Based on our findings, we propose 
{\em MI-resilient architecture designs} for the first time. 
Without bells and
whistles, we show in extensive experiments that our
MI-resilient architectures can outperform 
state-of-the-art (SOTA) defense methods in MI robustness.
Furthermore, 
our MI-resilient architectures are complementary to existing MI defense methods.
\textbf{Our project is available at \hyperlink{https://Pillowkoh.github.io/projects/RoLSS/}{https://Pillowkoh.github.io/projects/RoLSS/}}
  \keywords{Model Inversion \and Skip Connection} \and Model Inversion Resilient Architecture
\end{abstract}

\section{Introduction}
   
As deep neural networks (DNNs) see growing deployment across various applications like face recognition \cite{meng2021magface, guo2020learning, huang2020curricularface, schroff2015facenet, dufumier2021contrastive, yang2022towards, dippel2021towards, chang2020end, krishna2019speech} and healthcare \cite{dufumier2021contrastive, yang2022towards, dippel2021towards,muller2022radiological,luo2022pseudo,mishra2022data}, concerns about the privacy implications of DNNs are on the rise. Many DNNs are trained on private and sensitive datasets. There is an increasing concern of potential leakage of information of these private training samples through malicious exploitation of the model.

One particular privacy threat that has garnered growing attention is {\bf Model Inversion (MI) attacks}. In MI attacks, adversaries seek to reconstruct private training samples by exploiting vulnerabilities in the model. For instance, an adversary with access to a face recognition model may abuse it to reconstruct the private facial images of individuals from the model's training dataset. Following previous works \cite{zhang2020secret,nguyen_2023_CVPR,chen2021knowledge,han2023reinforcement}, we focus on reconstruction of images
and use the face recognition as a running example.

{\bf Research gap.}
Recently, there is an increasing interest to study MI and to understand the feasibility and extent of reconstructing private training samples from DNNs, from the MI attack and MI defense perspectives. {\em However, previous studies have overlooked DNN architecture, and there is a lack of study to understand how DNN architecture designs affect MI} (Tab. \ref{tab:previous}). In particular, MI has been formulated as an optimization problem to seek an image similar to that of an identity in the private training dataset. Commonly, the MI optimization 
\begin{wraptable}{r}{6cm}
\vspace{-1.1cm}
\caption{
There is a lack of study to understand how DNN architecture designs affect MI. Previous MI studies are DNN architecture-agnostic, focusing on MI objective, effect of regularizing MI objective, effect of distributional prior based on generative modelling, and regularization on the training objective of the target model.
Our work is a pioneer study to understand how DNN architectures affect MI attacks.
}
\label{tab:previous}
\setlength{\tabcolsep}{0.7em}
\begin{adjustbox}{width=0.48\columnwidth,center}
    \begin{tabular}{|l|c|c|c|c|c|}
    \hline 
    & \rotatebox[origin=c]{90}{\makecell{MI \\ objective}} 
     & \rotatebox[origin=c]{90} {\makecell{Effect of \\ regularizing \\MI objective}} 
     & \rotatebox[origin=c]{90} {
     \makecell{Effect of \\distributional \\prior to \\guide MI} } 
     & \rotatebox[origin=c]{90} {
     \makecell{Regularization \\ on the training \\ objective of \\target model} } 
     & \rotatebox[origin=c]{90} { \bf{\makecell{Effect of DNN \\ architecture \\ design on MI}} } 
     \\
    \hline
    \T\B MI~\cite{fredrikson2014privacy} &\checkmark & & & & \\
    \hline
    \T\B GMI~\cite{zhang2020secret} & &\checkmark &\checkmark & & \\
    \hline
    \T\B KEDMI~\cite{chen2021knowledge} & &\checkmark &\checkmark & & \\
    \hline
    \T\B VMI~\cite{wang2021variational} & & &\checkmark & & \\
    \hline
    \T\B MIRROR~\cite{an2022mirror} & & &\checkmark & & \\
    \hline
    \T\B PPA~\cite{struppek2022plug} &\checkmark & & & & \\
    \hline
    \T\B LOMMA~\cite{nguyen_2023_CVPR} &\checkmark & & & & \\
    \hline
    \T\B PLGMI~\cite{yuan2023pseudo} &\checkmark & &\checkmark & & \\
    \hline
    \T\B RLBMI~\cite{han2023reinforcement} &\checkmark & & & & \\
    \hline
    \T\B BREPMI~\cite{kahla2022label} &\checkmark & & & & \\
    \hline
    \T\B MID~\cite{wang2021improving} & & & &\checkmark & \\
    \hline
    \T\B BiDO~\cite{peng2022bilateral} & & & &\checkmark & \\
    \hline
    {\bf Ours} & & & & & \textbf{\checkmark}\\
    \hline
    \end{tabular}
\end{adjustbox}
\vspace{-0.5cm}
\end{wraptable}
objective is formulated as maximization of likelihood under the model being attacked (target model). 
Several improved MI objectives have been proposed recently, e.g. logit maximization \cite{nguyen_2023_CVPR,yuan2023pseudo}. Meanwhile, various regularizations on the MI objective have been studied to improve the effectiveness of MI, e.g. prior loss to penalize unrealistic images \cite{zhang2020secret}. In addition, various distributional priors leveraging generative models trained on public datasets have been proposed to guide the inversion (optimization) process during MI attacks \cite{chen2021knowledge,zhang2020secret,wang2021variational,an2022mirror,yuan2023pseudo}. Furthermore, regularizations on the training objective of the target model to reduce the correlation exploited by MI have been studied as methods to defend against MI attacks \cite{wang2021improving,peng2022bilateral}. However, there is a lack of study to understand the effect of DNN architecture design on MI.

\begin{figure}[ht!]%
  \centering
  \begin{adjustbox}{width=0.96\textwidth,center}
  \includegraphics[width=0.96\textwidth]{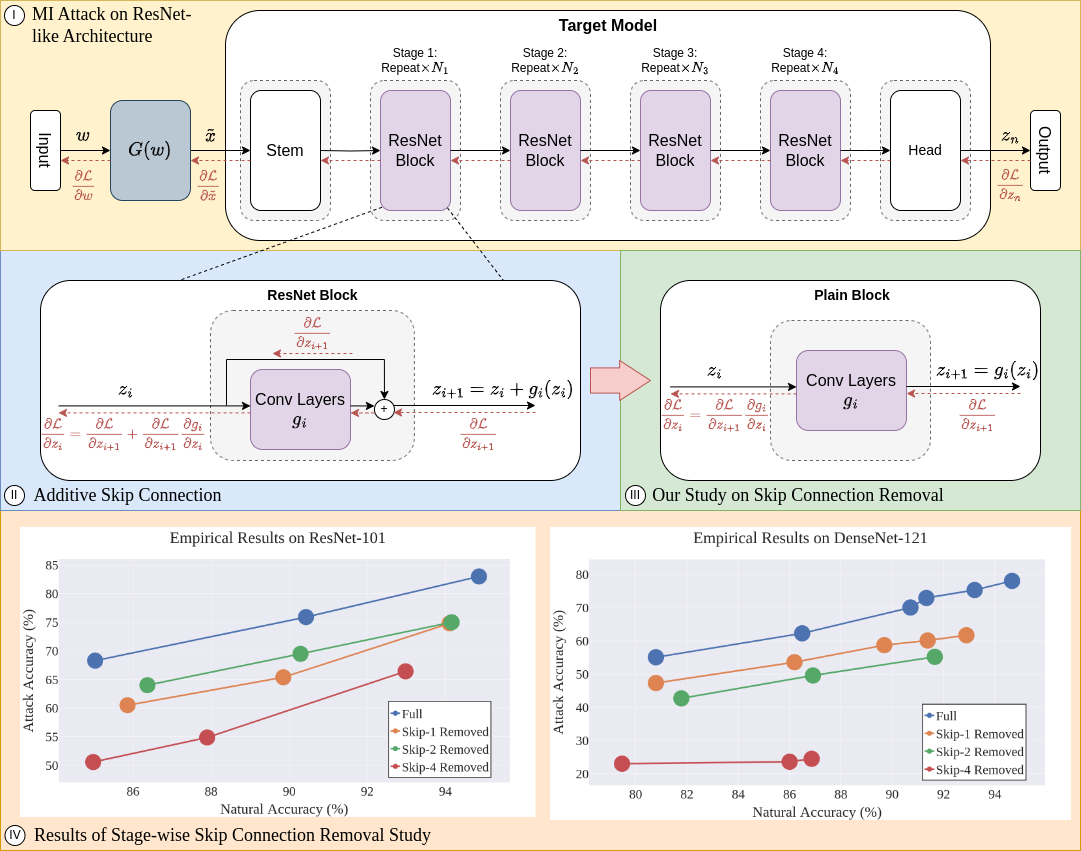}
  \vspace{-0.4cm}
  \end{adjustbox}
  \caption{
{\bf (I) Illustration of MI attack on ResNet-like architecture (Sec.~\ref{ssec:MI_intro})}.
This figure depicts the MI attack framework for SOTA white-box MI attacks \cite{zhang2020secret,chen2021knowledge,wang2021variational,yuan2023pseudo,nguyen_2023_CVPR,struppek2022plug,an2022mirror}, which leverage a generative model $G(.)$ to exploit the target model via gradient descent and backpropagation. Specifically, for each iteration, $  
\tilde{x} = G(w)$ is fed into the target model in the forward pass, and MI loss $\cL$ is computed. In the backward pass, gradients of $\cL$ are computed and back-propagated to obtain $\partial{\cL} / \partial{w}$, which is used to update $w$ to achieve reconstruction of private training data.
 {\bf(II) Additive Skip Connection (Sec.~\ref{ssec:MI_intro})}. During MI attacks, skip connections allow  gradients to bypass the residue module and enhance backpropagation. We hypothesize that this reinforces MI attacks.
 {\bf(III) Our study on skip connection removal (Sec.~\ref{ssec:removal} and Sec.~\ref{Sec:Extensive_Evaluation}).} To validate our hypothesis that skip connections could reinforce MI, we study the effect of skip connections on MI by removing  skip connections  within various stages of the target model. We study both  additive and concatenative skip connections.  
{\bf(IV) Results of stage-wise skip connection removal study (Sec.~\ref{ssec:removal} and Sec.~\ref{Sec:Extensive_Evaluation}).} The sub-figures show that skip connections have a considerable effect on MI. For both additive and concatenative skip connections, we observe that removal of  skip connections result in considerable degradation of MI attack accuracy.
Furthermore, we observe that {\em skip connections in the last stage have the most significant effect on MI.}
\textbf{Best viewed in color with zooming in.}
  }
  \label{fig:Architecture_Overview}
  \vspace{-1.0cm}
\end{figure}

{\bf In this paper}, we address this research gap and conduct the first study to understand how DNN architecture designs affect MI. We put our focus on skip connections \cite{he2016deep, huang2017densely}, a fundamental network design that facilitates the training of very deep neural networks. Skip connections mitigate the vanishing gradient problem during the training stage \cite{he2016deep}.
Meanwhile, many state-of-the-art (SOTA) MI attacks
\cite{zhang2020secret,nguyen_2023_CVPR,chen2021knowledge,struppek2022plug,yuan2023pseudo} require the use of gradients to guide the reconstruction of private training samples during the inversion stage. {\em We hypothesize that skip connections facilitate flowing of gradient during inversion, reinforcing  MI attacks and posing a considerable vulnerability to data privacy in DNNs} 
(Fig. \ref{fig:Architecture_Overview}).
To validate our hypothesis, we carefully design experiments to single out the effect of skip connections on MI attack performance. 
Our extensive experiments on SOTA networks (ResNet \cite{he2016deep}, DenseNet \cite{huang2017densely}, MaxViT \cite{tu2022maxvit}, EfficientNet \cite{tan2019efficientnet}) against SOTA MI attacks (PPA \cite{struppek2022plug}, PLG-MI \cite{yuan2023pseudo}, LOMMA \cite{nguyen_2023_CVPR}, KEDMI \cite{chen2021knowledge})
 consistently show that skip connections reinforce MI attacks in DNNs.
Furthermore, we find that {\em skip connections in the last stage have the most significant effect to MI attacks.}

To mitigate the MI vulnerability caused by skip connection, we analyze, RepVGG \cite{ding2021repvgg}, an established reparameterization technique which converts a multi-branch training-time architecture with skip connections to a plain, VGG-like inference-time architecture with skip connection removed. However, {\em our analysis shows that, while RepVGG could enable an inference-time network without any skip connection, the gradients during MI attacks on this inference-time network are the same as that on the multi-branch training-time architecture with skip connections.} Therefore, RepVGG could not mitigate the vulnerability.


To bridge the existing gap, we propose {\em MI-resilient architecture designs} based on our own findings. Specifically, 
as we find that the last stage's skip connections have the most significant effect to MI, we propose 
Removal of Last Stage Skip-Connection (RoLSS) as MI-resilient architecture designs.
As our designs remove only the last stage's skip connections and keep other stages' skip connections intact, we could keep the impact on natural accuracy small in many cases.
Building on top of RoLSS, we propose Skip-Connection Scaling Factor (SSF) and Two-Stage Training Scheme (TTS) to
recover the model's natural accuracy while maintaining competitive MI robustness.
Our contributions are:
\begin{itemize}
\item
We conduct a pioneer study to understand how  skip connections impact MI attacks and MI robustness.
We design experiments to carefully single out the effect of skip connections on MI attack performance, accounting for the effect of  natural accuracy in our analysis. Through extensive experiments spanning 4 SOTA MI attacks and 10 architectures, we validate that skip connections reinforce  MI attacks and pose a considerable vulnerability to data privacy in DNNs (Sec.~\ref{Sec:Ablation Study}).

\item Notably, we discover that {\em skip connections in the last stage consistently are the most critical to MI attacks} (Sec.~\ref{Sec:Ablation Study} and Sec.~\ref{Sec:Extensive_Evaluation}).

\item
We analyze RepVGG, a well-established reparameterization technique that remove skip connections by decoupling training-time and inference-time architectures. Our analysis reveals that this approach could not mitigate vulnerability to MI attacks
(Sec.~\ref{Sec:RepVGG}).

\item
Based on our findings, we propose MI resilient architecture designs for the first time, including: Removal of Last Stage Skip-Connection (RoLSS), Skip-Connection Scaling Factor (SSF) and Two-Stage Training Scheme (TTS). 
Extensive experiments show that 
our MI-resilient architectures can outperform
SOTA 
defense methods in MI robustness
(Sec.~\ref{Sec:Application}).

\end{itemize}




\section{Related Work}

{\bf Model Inversion.} 
The concept of MI was initially studied by Fredrikson et al. \cite{fredrikson2014privacy}, who demonstrated that adversaries could employ machine learning to extract genomic and demographic information about patients from a medical imaging model. This work was later extended to facial recognition in \cite{fredrikson2015model}.  
Since then, several MI studies have been conducted to understand the feasibility and extent of reconstructing private training samples from DNNs \cite{yang2019neural,wang2021variational,struppek2022plug,kahla2022label,nguyen_2023_CVPR,yuan2023pseudo,han2023reinforcement,zhang2020secret,wang2021improving,peng2022bilateral}.
These studies encompass both MI attacks and MI defense perspectives. We summarize notable developments in Tab.~\ref{tab:previous}. See Supp. for further discussion of related work.
{\em Despite considerable  progress in MI research, there is a lack of study to understand the effect of DNN architecture design on MI.}

\noindent {\bf Skip connections and DNNs Attacks.} Skip connections are recognized as an effective approach to alleviate the vanishing gradient problem, allowing us to train very deep neural networks \cite{he2016deep, huang2017densely}.
There are a few works that study the effect of skip connections to adversarial attacks \cite{wu2020skip, cazenavette2021architectural} and backdoor attacks \cite{yang2023backdoor}. \textbf{Differ from existing works, our study is the first to understand how DNN architectures affect model inversion (MI), a growing privacy attack.} Our investigation reveals a distinctive aspect: while previous work observed that  skip connections aids adversarial robustness \cite{cazenavette2021architectural},
our work instead discovers that 
skip connections reinforce MI attacks. It is important to emphasize that the nature of MI attacks differs significantly from adversarial or backdoor attacks. Particularly, for adversarial attacks, the goal is to deceive the model into making incorrect predictions. For backdoor attacks, the goal is to implant malicious functionality in the model such that the model produces incorrect outputs when a specific attack trigger is present in the input. Importantly, adversarial attacks/ backdoor attacks are not privacy attacks and they do not extract sensitive training data information from ML models. Rather, adversarial attacks/ backdoor attacks aim to undermine model utility and robustness.
{\em Our work is the first to study the implications of skip connection on data privacy of ML models through the MI attacks.}

\section{An Investigation on the Skip Connection Vulnerability to Model Inversion Attacks} \label{Sec:Ablation Study}

\subsection{Skip connections and MI attacks}
\label{ssec:MI_intro}
We first discuss the potential effect of skip connections on MI in this sub-section. Then, the effect is validated in  Sec. \ref{ssec:removal} and Sec.~\ref{Sec:Extensive_Evaluation}.

{\bf MI and gradients.}
MI attacks are a data privacy threat. 
For a 
DNN model $T$   trained on a private training dataset $\mathcal{D}_{priv}$, 
the adversary tries to exploit sensitive training data $\mathcal{D}_{priv}$ via the trained model $T$.
In most works, 
MI is formulated as the reconstruction of 
an input $\tilde{x}$ which is most likely classified into an identity   $y$ by $T$.
 The model $T$ subject to MI attacks is called {\em target model}.
We focus on {\em white-box} MI attack, which is the most popular and powerful MI attack
\cite{zhang2020secret,chen2021knowledge,nguyen_2023_CVPR,yuan2023pseudo,struppek2022plug,an2022mirror}.
Specifically, we follow previous works and assume attackers have access to the parameters, architectures, and outputs of the models \cite{zhang2020secret,chen2021knowledge,nguyen_2023_CVPR,yuan2023pseudo,struppek2022plug,an2022mirror}.

To reconstruct a  high-dimensional image $\tilde{x}$, some distributional priors have been proposed in SOTA MI to constrain the search space \cite{zhang2020secret,chen2021knowledge}.
The distributional prior is commonly encoded by a generative model $G(w)$ trained on 
a public dataset $\mathcal{D}_{pub}$ which has no class intersection with $\mathcal{D}_{priv}$.
MI attacks are commonly formulated as the following optimization:
\begin{equation}
\label{eqn:overall_mi_objective}
    w^*  = \arg \min_{w} 
    {\cL}(w;y,T) 
\end{equation}
Here,
${\cL}(w;y,T)$ is the MI loss which
 guides  reconstruction of $\tx=G(w)$ that is most likely to be classified by model $T$ as identity $y$.
Commonly, 
negative log-likelihood is used: 
${\cL}(w;y,T) = -\log \mathbb{P}_T(y|G(w))$, while other losses have been proposed, e.g., logit-based \cite{nguyen_2023_CVPR}.
In addition, other regularization can be included in $\cL$, e.g. prior loss \cite{zhang2020secret}.

Importantly, to solve the optimization in 
Eq. \ref{eqn:overall_mi_objective}, gradient descent and back propagation are used by most SOTA white-box MI attacks~\cite{zhang2020secret,chen2021knowledge,nguyen_2023_CVPR,yuan2023pseudo,struppek2022plug,an2022mirror}: For each iteration,  $G(w)$ is fed into $T$
in the forward pass, and $\cL$ is computed.
In the backward pass, gradients of $\cL$ are computed in $T$ and back-propagated to obtain 
$\partial{\cL} / \partial{w}$, which is used
to update $w$ by the attackers.

{\bf Skip connections could reinforce MI.}
Following the above discussion, backpropagation of gradients during MI inversion could have a considerable effect on the MI attack performance. Meanwhile, for conventional DNN training, skip connections are a fundamental architecture design that is effective in mitigating gradient vanishing during backpropagation. We hypothesize that skip connections could facilitate gradient backpropagation during MI attacks and reinforce MI, thereby compromising data privacy.

Specifically, in a ResNet-like architecture, there are multiple ResNet blocks. Each ResNet block, with input $z_i$ and output $z_{i+1}$, can be represented as: $z_{i+1} = z_i + g_i(z_i)$, including an additive skip connection and a residual module $g_i$ comprising multiple convolutions (Fig. \ref{fig:Architecture_Overview}).
During MI inversion, the gradient backpropagates across a ResNet block as follows:
{\small
\begin{equation}
    \pdv{\cL}{z_{i}} = \pdv{\cL}{z_{i+1}}\pdv{z_{i+1}}{z_i} 
            =\pdv{\cL}{z_{i+1}}\left(1+\pdv{g_i}{z_{i}}\right)
            = \textcolor{magenta}{\pdv{\cL}{z_{i+1}}} + \pdv{\cL}{z_{i+1}}\pdv{g_i}{z_{i}}
\label{eq:Skip_Connection_Gradient0}
\end{equation}
}
Importantly, 
the first gradient component, $\textcolor{magenta}{\pdv{\cL}{z_{i+1}}}$, 
enabled by the skip connection, enhances backpropagation. We hypothesize that this reinforces MI attacks.

\subsection{Stage-wise skip connection removal study}
\label{ssec:removal}
In this section, we validate  effect of skip connections on MI.

\noindent{\bf MI setup. } We conduct our analysis on ResNet-101 \cite{he2016deep} and DenseNet-121 \cite{huang2017densely} as target models under the attack setup of SOTA MI attack method PPA \cite{struppek2022plug}.
We strictly follow PPA MI setups,
where we use FaceScrub \cite{ng2014data} as private dataset, $\mathcal{D}_{priv}$ and attack all IDs as per PPA setup. Following previous SOTA MI works \cite{zhang2020secret,chen2021knowledge,wang2021variational,kahla2022label,struppek2022plug,yuan2023pseudo,an2022mirror,nguyen_2023_CVPR}, we adopt attack accuracy (AttAcc), measured using an evaluation model, as the primary metric for assessing MI performance. Attack accuracy is defined as the percentage of reconstructed images correctly identified by the evaluation model with respect to the target ID. Specific MI attack configuration can be found in the Supp.

We conduct our study by removing skip connections from various stages of the architecture. Each time, skip connections from a specific stage are removed, while those in the remaining stages remain unchanged. Each architecture with removed skip connections is trained using $\mathcal{D}_{priv}$ in
exactly the same settings as 
the original unaltered architecture. In this study, we focus on two common skip connection mechanisms: additive and concatenative.
 
\noindent {\bf Additive skip connection removal. } We investigate additive skip connections within individual stages of the ResNet-101 architecture. To remove additive skip connections in ResNet-like architectures, we ensure that outputs from the previous layers are not added to the subsequent layers during the feedforward process, as shown in Fig.~\ref{fig:Architecture_Overview}-III. For ResNet-101, the resulting ResNet block encompasses a convolutional layer, $g_i$, that consists of a single $1\times1$ convolution, followed by a $3\times3$ convolution, and lastly, another $1\times1$ convolution layer, without a shortcut connection linking the input and output of the ResNet block. 

\begin{table} [t]
\caption{We strictly follow SOTA PPA \cite{struppek2022plug} for the attack setup and evaluation. Here $\mathcal{D}_{priv}$ = Facescrub\cite{ng2014data}, $\mathcal{D}_{pub}$ = FFHQ \cite{karras2019style}. Across architectures both additive and concatenative skip connections, we consistently observe that {\bf Skip connections in the last stage are the most critical to MI attacks}, resulting in the most degradation in MI attack accuracy. $\Delta_{AttAcc}$ represents the reduction in attack accuracy when compared to ``Full'' setting.}
\label{tab:skip4_evidence}
\parbox{.47\linewidth}{
\centering
\vspace{-0.3cm}
\setlength{\columnsep}{1pt}
\setlength\tabcolsep{2pt}
\begin{adjustbox}{width=0.45\textwidth,center}

\begin{tabular}{cccc}
\hline
\multicolumn{1}{l}{Architecture} & \multicolumn{1}{l}{Skip Connections} & \multicolumn{1}{l}{AttAcc} & \multicolumn{1}{l}{$\Delta_{AttAcc}$} \\ \hline
\multirow{4}{*}{ResNet-34}                        & Full                                 & 90.78                      & -                                     \\
\multicolumn{1}{l}{}             & Skip-1 Removed                       & 83.25       &  7.53                                     \\
\multicolumn{1}{l}{}             & Skip-2 Removed                       & 77.92       &     \textbf{12.86}                                  \\
\multicolumn{1}{l}{}             & Skip-4 Removed                       & 80.61                      & 10.17                                 \\ \hline
\multirow{4}{*}{ResNet-50}       & Full                                 & 82.76                      & -                                     \\
                                 & Skip-1 Removed                       & 73.23                      & 9.53                                  \\
                                 & Skip-2 Removed                       & 78.77                      & 3.99                                  \\
                                 & Skip-4 Removed                       & 68.44                      & \textbf{14.32}                                 \\ \hline
\multirow{4}{*}{ResNet-101}      & Full                                 & 83.00                      & -                                     \\
                                 & Skip-1 Removed                       & 74.81                      & 8.19                                  \\
                                 & Skip-2 Removed                       & 78.75                      & 4.25                                  \\
                                 & Skip-4 Removed                       & 58.68                      & \textbf{24.32}                                 \\ \hline
\multirow{4}{*}{ResNet-152}      & Full                                 & 86.51                      & -                                     \\
                                 & Skip-1 Removed                       & 80.35       &    6.16                                   \\
                                 & Skip-2 Removed                       &  69.04      &        17.47                               \\
                                 & Skip-4 Removed                       & 68.44                      & \textbf{18.07}                                 \\ \hline
\end{tabular}
\end{adjustbox}
}
\hfill
\parbox{.47\linewidth}{
\vspace{-0.3cm}
\centering
\setlength{\columnsep}{1pt}
\setlength\tabcolsep{2pt}
\begin{adjustbox}{width=0.455\textwidth,center}
\begin{tabular}{cccc}
\hline
\multicolumn{1}{l}{Architecture} & \multicolumn{1}{l}{Skip Connections} & \multicolumn{1}{l}{AttAcc} & \multicolumn{1}{l}{$\Delta_{AttAcc}$} \\ \hline
\multirow{4}{*}{DenseNet-121}    & Full                                 & 88.11                      & -                                     \\
                                 & Skip-1 Removed                       & 61.72                      & 26.39                                 \\
                                 & Skip-2 Removed                       & 55.21                      & 32.90                                 \\
                                 & Skip-4 Removed                       & 24.48                      & \textbf{63.63}                                 \\ \hline
\multirow{4}{*}{DenseNet-161}    & Full                                 & 74.86                      & -                                     \\
                                 & Skip-1 Removed                       &  55.38      &    19.48                                   \\
                                 & Skip-2 Removed                       &  59.25      &      15.61                                 \\
                                 & Skip-4 Removed                       & 20.71                      & \textbf{54.15}                                 \\ \hline
\multirow{4}{*}{DenseNet-169}    & Full                                 & 77.15                      & -                                     \\
                                 & Skip-1 Removed                       & 60.99                      & 16.16                                 \\
                                 & Skip-2 Removed                       & 51.86                      & 25.29                                 \\
                                 & Skip-4 Removed                       & 6.77                       & \textbf{70.38}                                 \\ \hline
\multirow{4}{*}{DenseNet-201}    & Full                                 & 77.62                      & -                                     \\
                                 & Skip-1 Removed                       &  57.41      &        20.21                               \\
                                 & Skip-2 Removed                       &   46.65     &             31.97                          \\
                                 & Skip-4 Removed                       & 20.71                      & \textbf{56.91}                                 \\ \hline
\end{tabular}
\end{adjustbox}
}
\vspace{-2em}
\end{table}

\noindent {\bf Concatenative skip connection removal. } We investigate concatenative skip connections within individual stages of the DenseNet-121 architecture. We remove the concatenative skip connections in a similar manner as the removal of additive skip connections (The details can be found in the Supp). DenseNet architectures contain DenseBlocks where input features are concatenated with the output features, before being fed into the next DenseBlock. When these features are merged through concatenation, each layer has direct access to the gradients from the loss function and the original input image. To remove these concatenative skip connections, we remove the process of concatenation and only pass the output feature of the current DenseBlock to the subsequent DenseBlock. 

\noindent{\bf Skip connections reinforce MI.} To benchmark our experiments, we utilize the original unaltered architecture to assess the performance of the modified architectures. For a fair comparison, we consider the strong correlation between natural accuracy and attack accuracy \cite{zhang2020secret}. Thus, we compare these architectures at multiple checkpoints that achieve similar natural accuracy. When presenting our results, we denote the removal of all skip connections in the $N^{th}$ stage as ``Skip-N Removed''.  ResNet and DenseNet consist of {\bf four stages}. We examine removal of  skip connections from stages 1, 2, and 4. The result of removing skip connections in stage 3 are not included in our study, as this stage contains many parameters, and its removal leads to a severe reduction in model accuracy. 

Our findings reveal that architectures with fewer skip connections impede MI attacks, leading to a decrease in MI attack accuracy. As depicted in Fig.~\ref{fig:Architecture_Overview}-IV, both additive and concatenative skip connection studies consistently show that architectures labeled as ``Skip-N Removed'' exhibit low MI attack accuracy compared to the original architecture. 

\noindent{\bf Removal of Last Stage Skip-Connection (RoLSS) is the most critical to MI attacks.} Notably, we consistently observe that removing the skip connection in the last stage (i.e., ``Skip-4 Removed'') results in the most degradation in MI attack accuracy. We attribute this to the specific position of skip connections removed within the architecture. During gradient backpropagation in MI attack, gradients in earlier stages depend on those in later stages, as illustrated in Eq.~\ref{eq:Skip_Connection_Gradient0}. When skip connections in stage 4 (last stage) are removed, the degraded gradients in stage 4 permeate throughout the earlier stages of the architecture, resulting in the most degradation in MI attack accuracy. We further validate this observation on various architectures for both additive and concatenative skip connections in Tab.~\ref{tab:skip4_evidence}.

\begin{wrapfigure}{r}{0.5\textwidth}
\vspace{-0.6cm}
\centering
    \includegraphics[width=0.42\textwidth]{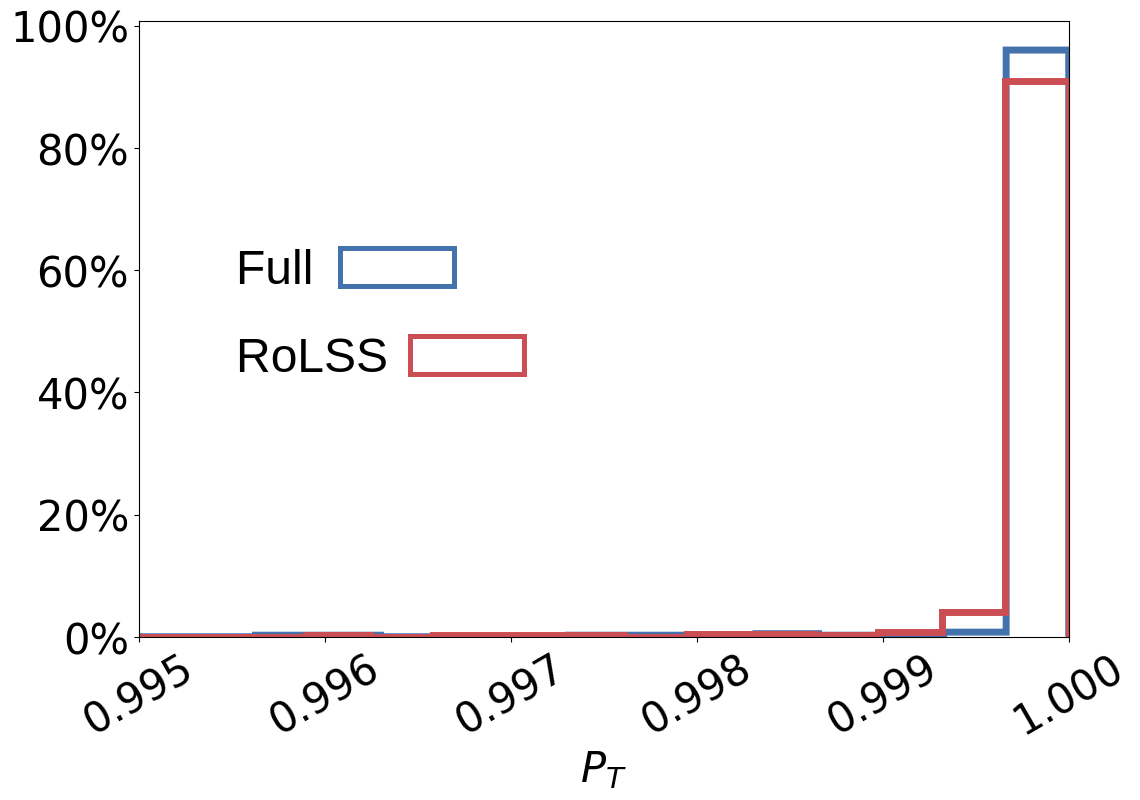}
    \caption{\textbf{MI  convergence analysis.} We compare histograms of likelihoods for  original architecture (Full) and Removal of Last Stage Skip-Connection (RoLSS) architecture for ResNet-101 under PPA attack.}
    \vspace{-0.6cm}
    \label{fig:analysis}
\end{wrapfigure}

\noindent{\bf Removing the skip connections leads to MI optimization converging with more false positives.} As discussed in Eq.~\ref{eqn:overall_mi_objective}, MI adversaries aim to identify optimal $w^*$ that maximizes the likelihood $\mathbb{P}_T(y|G(w))$. We provide this key observation to understand why removing skip connections degrades MI attacks: {\em When the skip connections are removed, latent variables with high likelihood $\mathbb{P}_T(y|G(w))$ can still be identified by Eq.~\ref{eqn:overall_mi_objective}, but many $w^*$ are false positives.} Consequently, this leads to a notable decrease in the accuracy of MI attacks. This observation becomes evident when examining the likelihood distribution of ``Full'' and ``Skip-4 removed'' settings (see Fig.~\ref{fig:analysis}), which are similar and both very close to 1. This results imply that, with ``Skip-4 removed'' setting, Eq.~\ref{eqn:overall_mi_objective} could still perform well to seek latent variables $w$ to maximize the likelihood $\mathbb{P}_T(y|G(w))$. However, despite the similarity in likelihood distributions, the attack accuracy of the ``Skip-4 removed'' setting is significantly lower than that of the ``Full'' setting (i.e., 24.32\%). This suggests that, due to the absence of skip connections, the gradients in the ``Skip-4 removed'' setting lead MI adversaries to exploit many $w^*$ that do not correspond to images resembling private data. 

\subsection{Extensive validation of the MI vulnerability of skip connections} \label{Sec:Extensive_Evaluation}

We conduct extensive experiments to further validate the impact of skip connections to MI attacks. Tab.~\ref{tab:Setup} summarize all MI setups in our validation ranging 10 architectures, 4 SOTA MI attacks, 3 datasets.

\begin{figure}[t]
  \centering
  \begin{adjustbox}{width=1.0\textwidth,center}
  \includegraphics[width=1.0\textwidth]{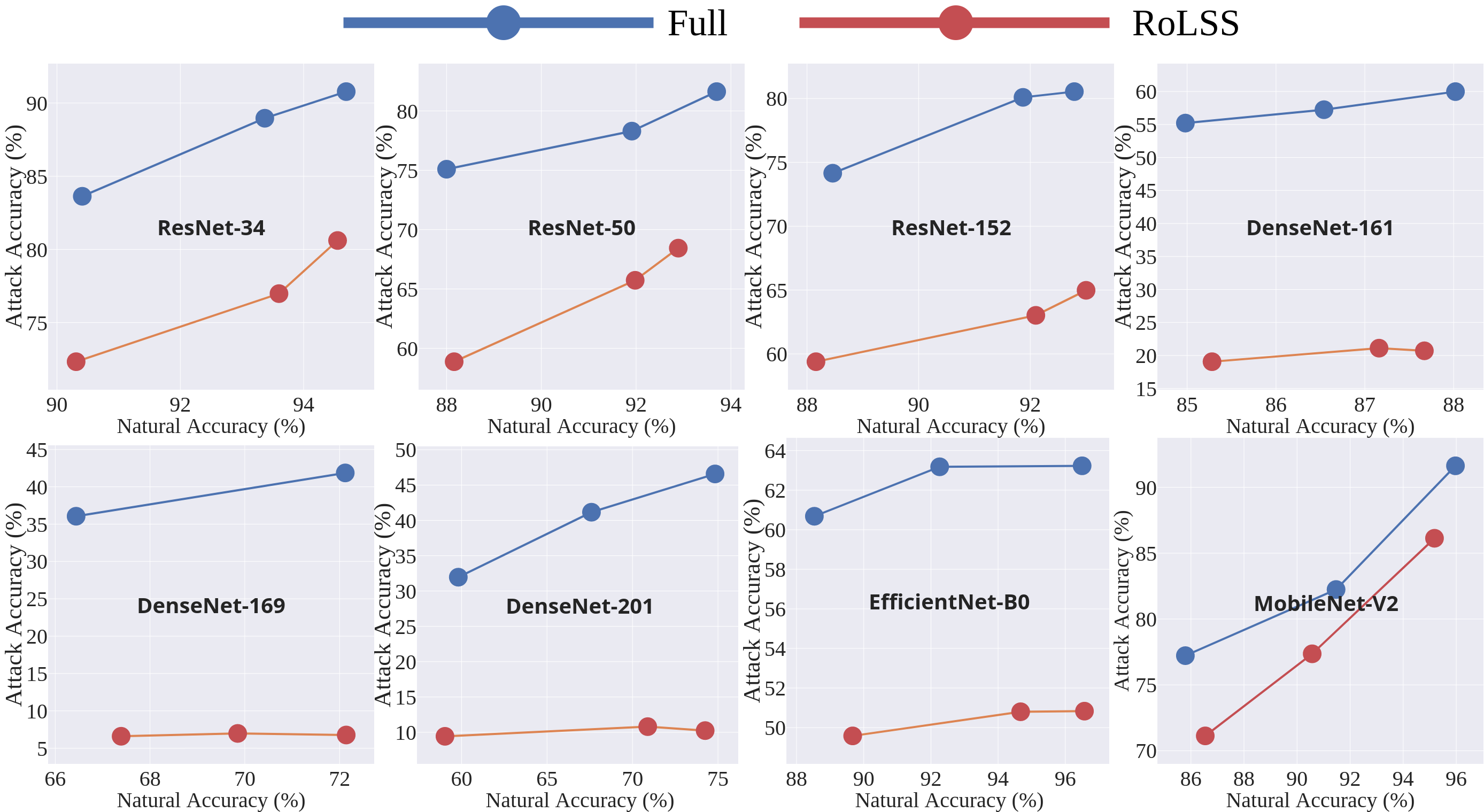}
  \end{adjustbox}
  \caption{
  Additional experiments to validate the impact of skip connections on MI attacks across various architecture designs, including networks with additive skip connections (ResNet-34/50/152 \cite{he2016deep}), concatenative skip connections (DenseNet-161/169/201 \cite{huang2017densely}), and compact CNN (EfficientNet-B0 \cite{tan2019efficientnet}).
  We strictly follow PPA \cite{struppek2022plug} for MI setups. In all cases, a significant decrease in attack accuracy is consistently observed when skip connections are removed in the {\em last stage}, demonstrating that skip connections reinforce MI attacks.
}
  \label{fig:Other_T}
  \vspace{-0.5cm}
\end{figure}

\noindent \textbf{Experimental Setting. } For a fair comparison, we follow the previous MI works \cite{zhang2020secret,nguyen_2023_CVPR,chen2021knowledge,han2023reinforcement,an2022mirror,struppek2022plug,yuan2023pseudo,peng2022bilateral,wang2021variational,kahla2022label} to select Evaluation Metrics, Private Dataset, Public Dataset, and Data Preparation Protocol. The details can be found in Tab.~\ref{tab:Setup}. Additional details are presented in the Supp.

\begin{wraptable}{r}{6cm}
\vspace{-0.9cm}
\caption{{\bf The summary of our MI setups.} We follow the exact the experiment setups of PPA \cite{struppek2022plug}. For the other MI attacks, we follow the setups in \cite{nguyen_2023_CVPR,chen2021knowledge} for \cite{chen2021knowledge,zhang2020secret,nguyen_2023_CVPR}. In total, our study spans 10 architectures, 4 MI attacks, 3 datasets.}
\label{tab:Setup}
    \begin{adjustbox}{width=0.49\columnwidth}
    \begin{tabular}{ccc}
    \hline
    \textbf{Architectures} \T\B   & \textbf{MI Attack}   & \textbf{Private Dataset}           \\ \hline
    ResNet-34/50/101/152 \cite{he2016deep} \T    & \multirow{6}{*}{PPA \cite{struppek2022plug}}  & \multirow{4}{*}{Facescrub \cite{ng2014data}}       \\
    DesnseNet-121/161/169/201 \cite{huang2017densely} &                      &                                                            \\
    MaxViT-T  \cite{tu2022maxvit}                &                      &                                                            \\
    EfficientNet-B0 \cite{tan2019efficientnet}    \B       &                      &                                                           \\ \cline{1-1} \cline{3-3}
    ResNet-50/101 \cite{he2016deep}   \T         &        & \multirow{2}{*}{Stanford Dogs \cite{KhoslaYaoJayadevaprakashFeiFei_FGVC2011}}                                           \\
    DenseNet-121/169 \cite{huang2017densely} \B         &                      &                                                           \\ \hline
    \multirow{3}{*}{IR152 \cite{he2016deep}}    & KEDMI \cite{chen2021knowledge}    \T            & \multirow{4}{*}{CelebA \cite{liu2015deep}}             \\
                              & LOMMA  \cite{nguyen_2023_CVPR}   \B           &                          \\ \cline{2-2} 
                              
                              & \T\B PLG-MI   \cite{yuan2023pseudo}                &                                                          \\\hline
    \end{tabular}
    \end{adjustbox}
\vspace{-0.8cm}
\end{wraptable}

\noindent \textit{Skip connections removal.} 
We apply our finding of Removal of Last Stage Skip-Connection (RoLSS) for various architectures. 



\noindent \textit{Evaluation Metrics.} Folloing the previous MI works, we adopt natural accuracy (Acc) and Attack Accuracy (AttAcc) as the main evaluation metrics. The detailed description and additional qualitative results are presented in the Supp.



\noindent \textbf{Experimental results on various architectures. }We note that in Fig.~\ref{fig:Other_T}, we focus on the range of natural accuracy where the two setups overlap, allowing us to observe changes in attack accuracy at the same natural accuracy level. The results are consistent with our empirical study in Sec.~\ref{Sec:Ablation Study}, where the RoLSS of additive skip connections (e.g., ResNet-34/50/152/EfficientNet-B0) reduce the attack accuracy by around 10\% to 15\% while the RoLSS of concatenative skip connections (e.g., DenseNet-161/169/201) reduce the attack accuracy by around 30\% to 35\%. Overall, in all cases including addictive connections (ResNet-34/50/152), concatenative skip connections (DenseNet-161/169/201), compact CNN (EfficientNet-B0), we consistently observe the significant drops in MI attack accuracy, ranging from ~10\% to ~35\%. This results further validate our findings in Sec.~\ref{Sec:Ablation Study} that skip connections reinforce MI attacks. More results can be found in the Supp.

\begin{figure}[t]
  \centering
  \begin{adjustbox}{width=0.9\textwidth,center}
  \includegraphics[width=0.9\textwidth]{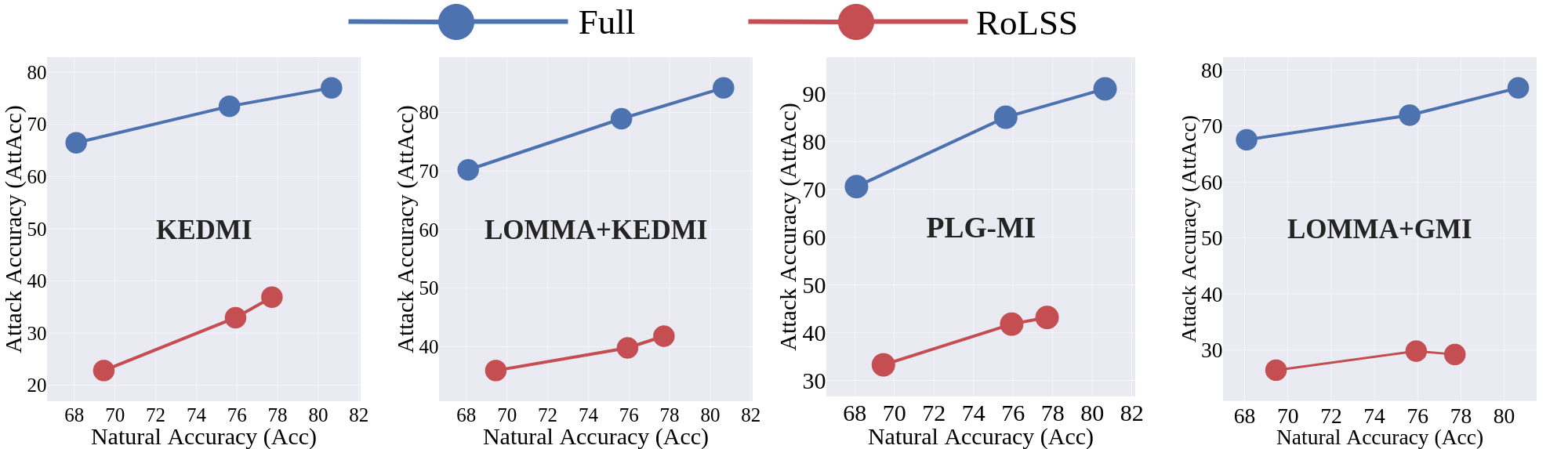}
  \end{adjustbox}
  \vspace{-0.6cm}
  \caption{\textbf{Additional experiments on other SOTA MI attacks} including KEDMI \cite{chen2021knowledge}, LOMMA \cite{nguyen_2023_CVPR}, and PLG-MI \cite{yuan2023pseudo}. We follow the standard setup, where $T$ = IR152, $\mathcal{D}_{priv}$ = CelebA, $\mathcal{D}_{pub}$ = CelebA/FFHQ. Across all SOTA MI attacks, a consistent and notable reduction in attack accuracy is observed when skip connections are removed in the last stage, demonstrating that skip connections reinforce MI attacks.}
  \label{fig:Other_Attack}
  \vspace{-0.5cm}
\end{figure}

\noindent \textbf{Experimental results on other SOTA MI attacks. }Beside PPA attack \cite{struppek2022plug}, we validate our findings of RoLSS on other SOTA MI attacks including KEDMI \cite{chen2021knowledge}, PLG-MI \cite{yuan2023pseudo} and LOMMA \cite{nguyen_2023_CVPR}. The details for these MI attack can be found in the Tab.~\ref{tab:Setup}. Following previous setups, we use IR152 \cite{he2016deep} as the architecture of the target classifier. The experimental results are presented in Fig.~\ref{fig:Other_Attack}. In Fig.~\ref{fig:Other_Attack}, we focus on the range of natural accuracy where the two setups overlap, allowing us to observe changes in attack accuracy at the same natural accuracy level. Across all these MI attacks, the results are consistent to those under PPA attack in Sec.~\ref{ssec:removal}, where the attack accuracy reduces significantly when the skip connections are removed regardless the effect from the natural accuracy.



\begin{figure}[t]
  \centering
  \begin{adjustbox}{width=0.9\textwidth,center}
  \includegraphics[width=0.9\textwidth]{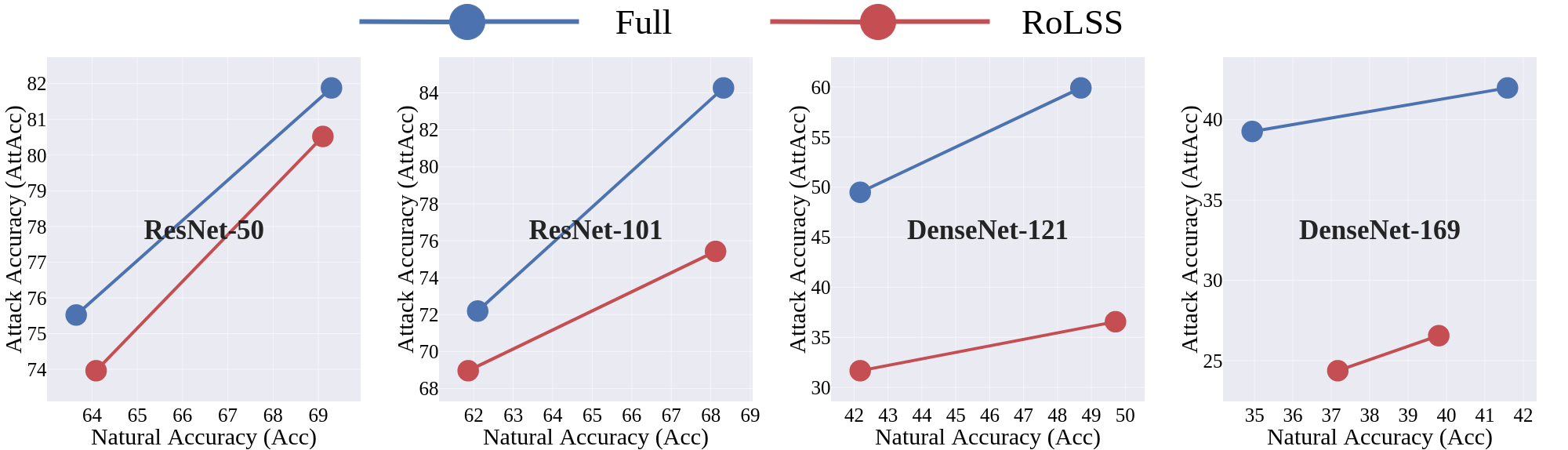}
  \end{adjustbox}
  \vspace{-0.5cm}
  \caption{
    \textbf{Additional experiments on the Stanford Dogs} \cite{KhoslaYaoJayadevaprakashFeiFei_FGVC2011} dataset as $\mathcal{D}_{priv}$.  The experiments are conducted under PPA \cite{struppek2022plug} attacks across various architectures, including ResNet-50/101 and DenseNet-121/169. We strictly follow MI setups in PPA.}
  \label{fig:StanfordDogs}
  \vspace{-0.6cm}
\end{figure}

\noindent \textbf{Experimental results on other private datasets. } In addion to Facescrub dataset \cite{ng2014data} in the main study, we further validate our findings of RoLSS on other datasets, including CelebA \cite{liu2015deep} and Stanford Dogs \cite{KhoslaYaoJayadevaprakashFeiFei_FGVC2011}. For CelebA \cite{liu2015deep}, we conduct our study on KEDMI \cite{chen2021knowledge}/LOMMA \cite{nguyen_2023_CVPR}/PLG-MI \cite{yuan2023pseudo}. We use the standard setup of IR152 \cite{he2016deep} as the architecture of target classifier. For Stanford Dogs, we conduct our study on PPA \cite{struppek2022plug}. We use the setup of ResNet-50/101 \cite{he2016deep} and DenseNet-121/169 \cite{huang2017densely} as the architecture of the target classifier. The experimental results are presented in Fig.~\ref{fig:Other_Attack} and Fig.~\ref{fig:StanfordDogs} for CelebA and Stanford Dogs, respectively. The results in both datasets are consistent with the results for Facescrub, where the attack accuracy experiences a significantly reduction when the skip connection is removed regardless the effect from the natural accuracy.


\section{Removing skip connection in inference-time architecture via RepVGG could not help mitigate vulnerability to MI} \label{Sec:RepVGG}

\begin{wrapfigure}{r}{0.5\textwidth}
\vspace{-1.5cm}
\centering
    \includegraphics[width=0.42\textwidth]{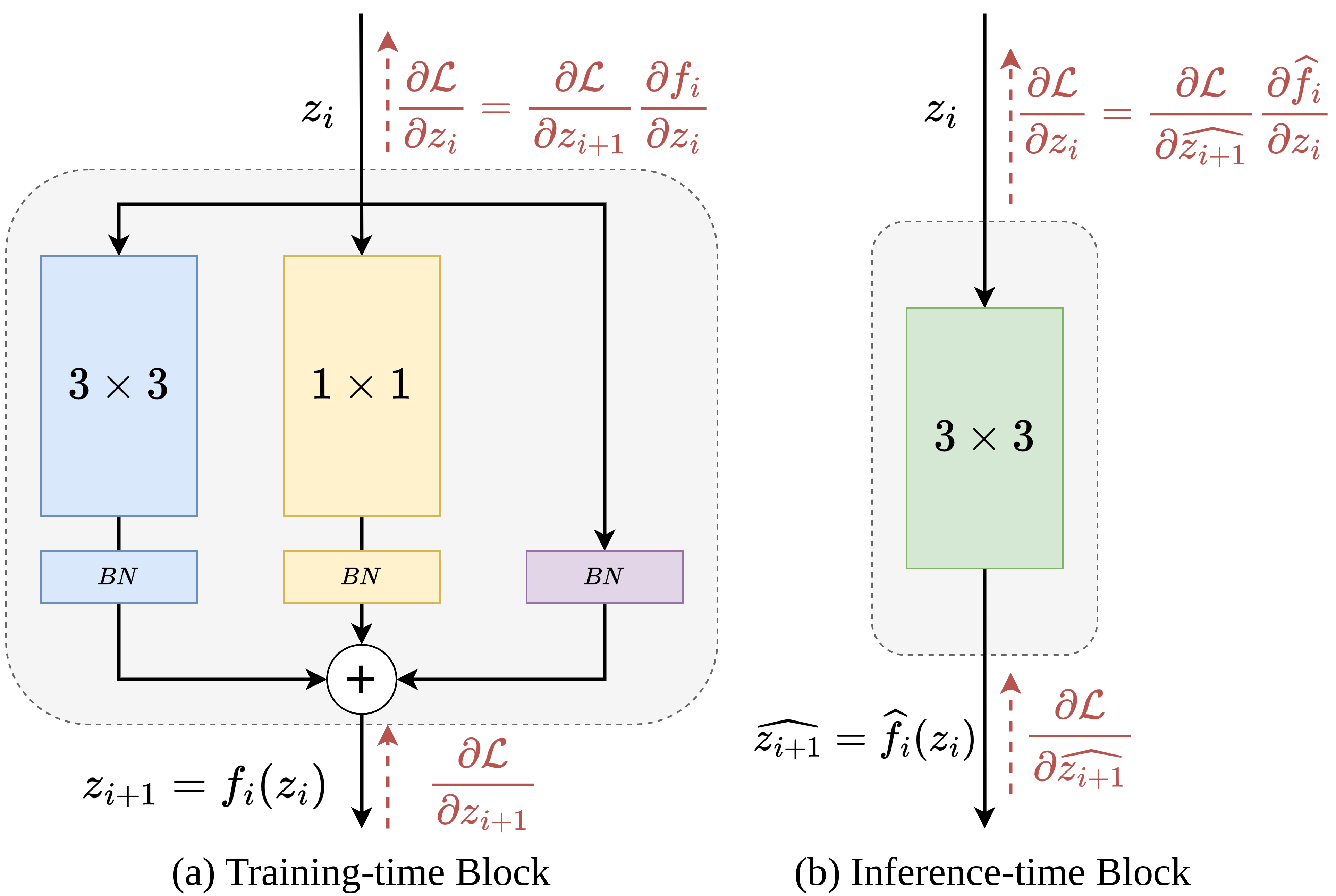}
    \caption{
      RepVGG \cite{ding2021repvgg} converts a training-time multi-branch  block (with skip connection) into an inference-time plain convolutional layer (without skip connection). Through our analytical and empirical analysis, we show that despite the differences in architectures,
      the gradients across the blocks are similar (See our analysis in the Supp.). Consequently, even with the removal of skip connections, RepVGG cannot mitigate vulnerability to MI attacks.
    }
    \vspace{-0.8cm}
    \label{fig:RepVGG_overview}
\end{wrapfigure}

In this section, we analyze RepVGG \cite{ding2021repvgg}, an established method to  decouple the training-time and inference-time architectures through structural re-parameterization. RepVGG converts a training-time multi-branch  block (with skip connection) into an inference-time plain convolutional layer (without skip connection) to accelerate inference speed, as shown in Fig.~\ref{fig:RepVGG_overview}. As RepVGG removes skip connections in  inference-time architecture, we seek to explore: \textit{Can RepVGG inference-time architecture mitigate vulnerability to MI attacks?}

We denote the training-time multi-branch architecture by $T_{RepVGG}$, and the inference-time plain architecture by $\widehat{T_{RepVGG}}$. Through our analytical and empirical analysis, we show that \textbf{the gradients in ${\bm{\widehat{T_{RepVGG}}}}$ under MI attacks and that in $\bm{T_{RepVGG}}$ are the same. Therefore, removing skip connection in the inference-time architecture 
$\bm{\widehat{T_{RepVGG}}}$ could not help mitigate vulnerability to MI.} 
Our detailed analytical analysis can be found in the Supp.

To empirically validate this,  we assess the vulnerability of RepVGG inference-time architecture to MI attacks, comparing it to the training-time architecture. We apply SOTA MI attack, PPA \cite{struppek2022plug}, on both the training-time and inference-time RepVGG-A0/B3/D2 architectures. The results for other RepVGG architectures can be found in Supp. We strictly follow the training and inference conversion implementation from the original RepVGG source code for the target classifiers trained on the Facescrub dataset \cite{ng2014data}. For the PPA attack \cite{struppek2022plug}, we follow the attack setup provided in the original PPA source code. 
Our results in Tab.~\ref{tab:RepVGG-PPA} clearly show that, \textbf{despite the removal of skip connections, the inference-time architecture obtained via RepVGG remains as vulnerable to MI attacks as the training-time architecture}.

\begin{wraptable}{r}{6cm}
\vspace{-1.2cm}
\caption{Experimental results of RepVGG \cite{ding2021repvgg} training-time and inference-time architectures. We strictly follow PPA \cite{struppek2022plug} for the attack setup and evaluation. Here $\mathcal{D}_{priv}$ = Facescrub, $\mathcal{D}_{pub}$ = FFHQ. Despite the removal of skip connections, RepVGG inference-time architecture  remains  as vulnerable to MI attacks as the training-time architecture.}
\label{tab:RepVGG-PPA}
  \begin{adjustbox}{width=0.45\columnwidth,center}
  \centering
    \begin{tabular}{cccc}
    \hline
                               & \textbf{Architecture} \T\B   & \textbf{Acc$\Uparrow$}   & \textbf{AttAcc$\Downarrow$} \\
                               \hline
    \multirow{2}{*}{RepVGG-A0} & Training-time \T\B  & 94.90 & 85.19 \\
                               & Inference-time \T\B & 94.90 & 86.13 \\
                               \hline
    \multirow{2}{*}{RepVGG-B3} & Training-time \T\B  & 94.55 & 80.69 \\
                               & Inference-time \T\B & 94.55 & 80.61 \\
                               \hline
    \multirow{2}{*}{RepVGG-D2} & Training-time \T\B  & 92.05 & 67.08 \\
                               & Inference-time \T\B & 92.05 & 66.44 \\
                               \hline
    \end{tabular}      
  \end{adjustbox}
\vspace{-0.7cm}
\end{wraptable}

The skip connection removed architectures in Sec.~\ref{Sec:Ablation Study} and those in the study of RepVGG in this section are fundamentally different. During training, RepVGG training-time architecture still receives gradients via skip connection branches. We show that the gradients in RepVGG inference-time architecture remain the same, including the skip connection branch's gradient components even though skip connections are absent in the inference-time architecture. In contrast, our study in Sec.~\ref{Sec:Ablation Study} removes skip connections during training, eliminating gradients on skip connections. The study demonstrates the causal effect of skip connections on MI attack accuracy.

\section{Model Inversion Resilient Architecture Design} \label{Sec:Application}

\begin{wraptable}{r}{6cm}
\vspace{-1.6cm}
\caption{{\bf Our simple RoLSS outperforms SOTA MI defense BiDO \cite{peng2022bilateral}. Our further proposed SSF and TTS help recover Acc while offer competitive MI robustness}. $\Delta$ represents the ratio of attack accuracy drop to natural accuracy drop. 
We could not compare with unsupported BiDO architectures (i.e., DenseNet), as BiDO requires extensive hyperparameter grid search.
}
\label{tab:MI-resilient-resnet}
  \begin{adjustbox}{width=0.45\columnwidth,center}
  \centering
    \begin{tabular}[b]{ccccc}
    \hline
    \textbf{Architecture}                            & \textbf{Defense} & \textbf{Acc$\Uparrow$ }         & \textbf{AttAcc$\Downarrow$}      & \textbf{$\Delta\Uparrow$ }         \\ \midrule
    \multirow{5}{*}{ResNet-34}                       & No Def.          & 94.69                & 90.78                & -   \\
                                                     & BiDO             & 91.66                & 81.98                & 2.90    \\
                                                     & RoLSS (Ours)           & 91.38                & 71.86                & 5.72   \\ 
                                                     & SSF (Ours)  &  94.21 & 79.79 & 22.90 \\
                                                     & TTS (Ours)  &  94.40 & 81.65 & \textbf{31.48} \\
                                                     \midrule
    \multirow{5}{*}{ResNet-50}                       & No Def.          & 94.58                & 82.76                & -   \\
                                                     & BiDO             & 91.12                & 58.41                & 7.04    \\
                                                     & RoLSS (Ours)           & 92.89                & 68.44                & \textbf{8.47}   \\
                                                     & SSF (Ours)  &  93.05 & 74.79 & 6.87 \\
                                                     & TTS (Ours)  &  93.56 & 77.21 & 5.44 \\
                                                     \midrule
    \multirow{5}{*}{ResNet-101}                      & No Def.          & 94.86                & 83.00                & -      \\            
                                                     & BiDO             & 90.31                & 67.07                & 3.50             \\
                                                     & RoLSS (Ours)           & 92.40                & 58.68                & 9.89  \\
                                                     & SSF (Ours)  &  93.79 & 71.06 & \textbf{11.16} \\
                                                     & TTS (Ours)  &  94.16 & 77.26 & 8.20 \\
                                                     \midrule
    \multirow{5}{*}{ResNet-152}                      & No Def.          & 95.43                & 86.51                & -   \\
                                                     & BiDO             & 91.80                & 58.14                & 7.82     \\
                                                     & RoLSS (Ours)           & 93.00                & 64.98                & 8.86  \\ 
                                                     & SSF (Ours)  &  93.79 & 70.71 & \textbf{9.63} \\
                                                     & TTS (Ours)  & 93.97  & 73.59 & 8.85 \\
                                                     \hline
    
    \end{tabular}    
  \end{adjustbox}
\vspace{-2.0cm}
\end{wraptable}

Our findings so far reveal that skip-connections reinforce MI attacks while existing reparameterization technique, RepVGG \cite{ding2021repvgg}, to remove skip connection in the inference-time architecture cannot mitigate vulnerability to MI attack. To bridge this gap, we propose MI-resilient architecture designs for the first time, including: Removal of Last Stage Skip-Connection (RoLSS), Skip-Connection Scaling Factor (SSF), and Two-stage Training Scheme (TTS). Our RoLSS, SSF, and TTS are remarkably simple, maintaining the same training procedure as the original model. Applying RoLSS involves no additional hyperparameters, while SSF/TTS requires only one extra hyperparameter, making them easily applicable to various architectures. In contrast, the SOTA MI defense BiDO requires an extensive grid search for each architecture \cite{peng2022bilateral}. 

\subsection{Removal of Last Stage Skip-Connection (RoLSS)}

In Sec.~\ref{Sec:Ablation Study}, our investigation reveals that removing skip connections in the last stage yields significant degradation in MI attack accuracy, suggesting a promising approach to improve MI robustness from architectural perspective.
The MI defense results are presented in Tab.~\ref{tab:MI-resilient-densenet} and Tab.~\ref{tab:MI-resilient-resnet}. Across architectures and skip connection mechanism, the results consistently show that \textbf{removing the skip connections in last stage (i.e., RoLSS) can improve the MI robustness.} 
Notably, our simple RoLSS achieves highly competitive MI robustness compared to the SOTA MI defense BiDO \cite{peng2022bilateral}. For instance, with ResNet-101, our RoLSS improves model accuracy by 2.09\%, while the MI attack accuracy degrades by 8.49\%, resulting in superior MI robustness compared to BiDO.


\subsection{Skip-Connection Scaling Factor (SSF)}




We further propose SSF on top of RoLSS to improve natural accuracy of the model while maintaining competitive MI robustness. 
For additive skip connections, we introduce  a scale factor $0 \leq k \leq 1$
for the signal on the skip connection of the last stage:
\begin{equation}
    z_{i+1} = g_i(z_i) + k \cdot z_i
    \label{Eq:SF_add}
\end{equation}

\begin{wraptable}{r}{6cm}
\vspace{-1.0cm}
\caption{{\bf Our simple RoLSS outperforms SOTA MI defense BiDO \cite{peng2022bilateral}. Our further proposed SSF and TTS help recover Acc while offer competitive MI robustness}. $\Delta$ represents the ratio of attack accuracy drop to natural accuracy drop. We could not compare with unsupported BiDO architectures (i.e., DenseNet), as BiDO requires extensive hyperparameter grid search.}
\label{tab:MI-resilient-densenet}
  \begin{adjustbox}{width=0.45\columnwidth,center}
  \centering
    \begin{tabular}[b]{ccccc}
    \hline
    \textbf{Architecture}                            & \textbf{Defense} & \textbf{Acc$\Uparrow$ }         & \textbf{AttAcc$\Downarrow$}      & \textbf{$\Delta\Uparrow$ }         \\ \midrule
    \multicolumn{1}{c}{\multirow{2}{*}{DenseNet-121}} & No Def.          & 94.67                & 78.09                & -    \\               
                                                     & RoLSS (Ours)           & 86.86                & 24.48                & 6.86    \\ 
                                                     & SSF (Ours)          &     91.73            &     56.32            & \textbf{7.35}   \\ \midrule
    \multicolumn{1}{c}{\multirow{2}{*}{DenseNet-161}} & No Def.          & 93.93                & 74.86                & -    \\
                                                     & RoLSS (Ours)           & 87.67                & 20.71                & \textbf{8.65}    \\
                                                     & SSF (Ours)          &   93.77              &      74.27           & 3.69   \\ \midrule
    \multicolumn{1}{c}{\multirow{2}{*}{DenseNet-169}} & No Def.          & 94.28                & 77.15                & -     \\                      
                                                     & RoLSS (Ours)           & 72.14                & 6.77                 & 3.18    \\ 
                                                     & SSF (Ours)          &        92.95         &     60.99            & \textbf{12.15}   \\ \midrule
    \multicolumn{1}{c}{\multirow{3}{*}{DenseNet-201}} & No Def.          & 94.32                & 77.62                & -    \\
                                                     & RoLSS (Ours)           & 74.25                & 10.24                & 3.36    \\
                                                     & SSF (Ours)          &      93.09           &       65.21          & \textbf{10.09}   \\ \hline
    \end{tabular}   
  \end{adjustbox}
\vspace{-0.7cm}
\end{wraptable}

Details of SSF for concatenative skip connections can be found in Supp.
Our SSF generalizes the skip connection, where $k=1$ corresponds to the original skip connection, while $k=0$ is similar to our skip connection removal study. 
With $k < 1$, gradients can be limited during MI attack, and this could degrade MI.

The results are presented in Tab.~\ref{tab:MI-resilient-densenet} and Tab.~\ref{tab:MI-resilient-resnet}. We apply SSF over RoLSS and set $k=0.01$ for all concatenative skip connection architectures and $k=0.2$ for all additive skip connection architectures in our experimental setups. Overall, our SSF  further improves MI robustness beyond RoLSS and outperforms the SOTA MI defense BiDO \cite{peng2022bilateral} across various architectures. Notably, for DenseNet, 
SSF significantly aids in recovering model performance while still mitigating MI attacks, resulting in much improved MI robustness. For example, with DenseNet-201, SSF only incurs a $\sim$1\% drop in model accuracy while degrading MI attack accuracy by  $\sim$12\%.

\subsection{Two-stage Training Scheme (TTS)}

We further introduce a Two-stage Training Scheme (TTS) on top of RoLSS to improve model accuracy while still maintaining competitive MI robustness. Inspired by Transfer Learning literature \cite{yosinski2014transferable}, TTS consists of two training stages:

\textit{Stage 1}: We train model $T$ with {\em full skip-connections architecture} over $M$ epochs. 
This stage ensures the reasonable convergence of $\theta_{T}$ with well-backpropagated gradients through the full skip connections architecture.
Note that model parameters are far from optimum initially. With full skip connections in this stage, large gradients can be backpropagated in making larger parameter updates.

\textit{Stage 2}: We remove skip connections in the last stage, i.e. RoLSS, to 
create {\em skip connection-removed architecture $T_{p}$}. Then, we continue to train  $\theta_{T_{p}}$ over $N$ epochs. The pre-trained parameters in Stage 1 serves as  initialization for $\theta_{T_p}$, thereby aiding the enhanced convergence of $T_p$.


We build TTS on top of RoLSS. For a fair comparison, the total training epochs for both stages (i.e., $M+N$) match the total training epochs of the original model. Across all setups, we set $M=5$ and $N=95$. The results in Tab.~\ref{tab:MI-resilient-resnet} demonstrate that TTS outperforms the SOTA MI defense BiDO \cite{peng2022bilateral}. Furthermore, TTS improves model accuracy while maintaining competitive MI robustness when compared to our RoLSS. For instance, in the ResNet-34 setup, TTS achieves similar MI attack accuracy as BiDO but maintains comparable model accuracy with No Def. model, achieving very competitive MI robustness.
\section{Conclusion}



We conducted  a pioneering study to examine the impact of DNN architecture on SOTA MI attacks. Our findings reveal that skip connections reinforce MI attacks, thereby jeopardizing data privacy. Through extensive MI setups, we find that the skip connections in the last stage is the most critical to MI attacks. Furthermore, our analytical and empirical analysis on RepVGG reveal that the removal of skip connections in the inference-time architecture could not help mitigate the MI vulnerability. Based on our own findings, we propose MI-resilient architecture designs for the first time, including: Removal of Last Stage Skip-Connection (RoLSS), Skip-Connection Scaling Factor (SSF), and Two-stage Training Scheme (TTS). Our MI-resilient architecture designs are remarkably simple to apply and achieve very competitive MI robustness compared to SOTA MI defense.


\section{Acknowledgement}

This research is supported by the National Research Foundation, Singapore under its AI Singapore Programmes (AISG Award No.: AISG2-TC-2022-007); The Agency for Science, Technology and Research (A*STAR) under its MTC Programmatic Funds (Grant No. M23L7b0021). This material is based on the research/work support in part by the Changi General Hospital and Singapore University of Technology and Design, under the HealthTech Innovation Fund (HTIF Award No. CGH-SUTD-2021-004).

\bibliographystyle{splncs04}
\bibliography{main}

\clearpage


\section*{Overview}

We provide additional results and analysis in this Supp, including:

\begin{itemize}
    \item Our skip connection removal study on Vision Transformers (Sec.~\ref{Sec:ViT}).
    \item  Detailed analysis and additional empirical results for RepVGG \cite{ding2021repvgg} under MI study (Sec.~\ref{Sec:Addtional RepVGG}).
    \item Additional discussion and results for our MI-resilient architectures (Sec.~\ref{Sec:Additional_Ours}).
    \item User study (Sec.~\ref{Sec:User Study}).
    \item Discussion on architectures without skip connection (Sec.~\ref{Sec:VGG})
    \item The detailed experimental setting for skip connection removal (Sec.~\ref{training setting}) and MI attack (Sec.~\ref{Sec:MI attack}).
    \item Further discussion on related works (Sec.~\ref{Sec:Related Work}).
    \item The limitation (Sec.~\ref{Sec:Limitation}) and ethical consideration (Sec.~\ref{Sec:Ethical Consideration}) of our work.
\end{itemize}


    




\section{Skip Connection Removal Study on Vision Transformer} \label{Sec:ViT}

Similar to the study on CNNs architectures in the main manuscript, we conduct the skip connection removal study on Vision Transformer (ViT) architectures. Specifically, we put our focus on vanila ViT \cite{dosovitskiy2020image} and MaxViT \cite{tu2022maxvit}. \textbf{Our observations are consistent with those in CNNs architectures, where skip connections reinforce MI attacks.}

\noindent \textit{MI Attacks.} To assess MI vulnerability, we employ the SOTA MI attack, PPA \cite{struppek2022plug}, utilizing StyleGAN \cite{karras2019style} as the prior distribution. In PPA, the attack is performed on StyleGAN $\mathcal{W}$ space, which is previously optimized from StyleGAN $\mathcal{Z}$ space during MI initialization stage.

\noindent \textit{Skip connections removal.} Due to  feature collapse phenomenon in ViTs:  Removing skip connections in the later stages \cite{tang2021augmented} results in very poor model performance for ViTs, we are only able to remove the skip connections in the first stage. 

\noindent \textit{Target classifier $T$.} We conduct our study on vanila ViT \cite{dosovitskiy2020image} and SOTA Vision Transformer architecture, MaxViT \cite{tu2022maxvit}.

\noindent \textit{Evaluation Metrics.} Folloing the previous MI works, we adopt natural accuracy (Acc) and Attack Accuracy (AttAcc) as the main evaluation metrics. 

\noindent \textit{Dataset ${\mathcal{D}_{priv}}$.} As facial recognition are commonly used in real-world scenarios, following the existing MI works, we focus on the study of Facescrub \cite{ng2014data}.

\noindent \textit{Data Preparation Protocol.} Following previous MI works, the private dataset $\mathcal{D}_{priv}$ is exclusively used for training the target classifier $T$, while the public dataset $\mathcal{D}_{pub}$ is utilized to extract prior information. There is no class intersection between $\mathcal{D}_{priv}$ and $\mathcal{D}_{pub}$ to ensure that the adversary only access to $\mathcal{D}_{pub}$ to extract general features, and does not access to the information about $\mathcal{D}_{priv}$ used for training target model. 

\begin{figure*}[h!]
  \centering
  \begin{adjustbox}{width=0.7\textwidth,center}
  \includegraphics[width=0.7\textwidth]{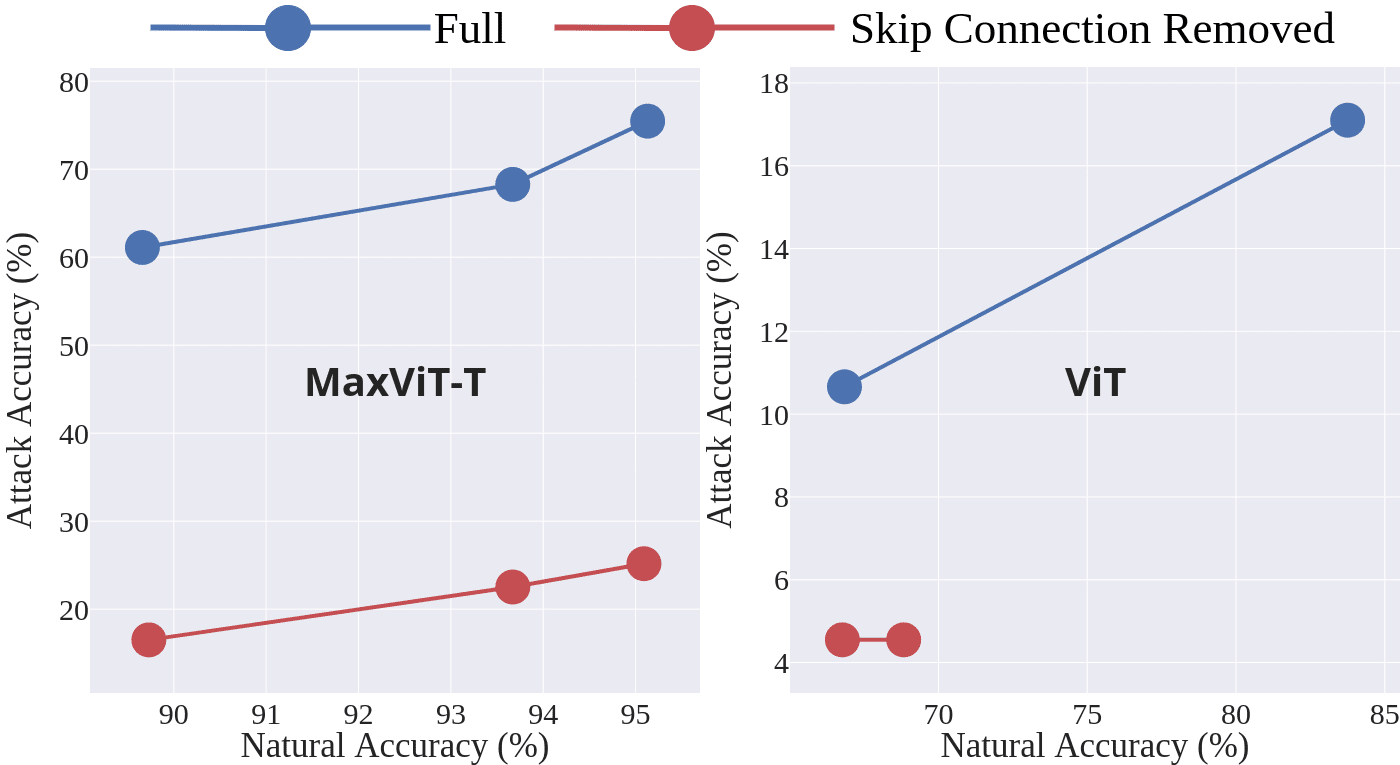}
  \end{adjustbox}
  \vspace{-0.2cm}
  \caption{{\bf Our Skip Connection Removal Study on Vision Transformer Architectures}. We follow the MI setup from PPA \cite{struppek2022plug}, $\mathcal{D}_{priv}$ = Facescrub, $\mathcal{D}_{pub}$ = FFHQ. Consistent with our observations in the CNNs, we note a significant decrease in attack accuracy when skip connections are removed, indicating that skip connections reinforce MI attacks.}
  \label{fig:ViT}
  \vspace{-0.2cm}
\end{figure*}

\noindent \textbf{Experimental results. }The results in Fig.~\ref{fig:ViT} are consistent with our study conducted on CNN architectures, where the skip connection reinfoce MI attack. For example, we observe a significant drop in MI attack accuracy $\sim$30\% in MaViT when skip connections are removed regardless the effect of natural accuracy. This results further validate our findings in the main manuscript.

Notably, in the MaxViT, the decrease in natural accuracy is minimal when skip connections are removed in a single stage, but the decrease in attack accuracy is significant. This leads to the improvement in MI robustness. As depicted in Fig.~\ref{fig:Viz},  the removal of skip connections in the MaxViT setup significantly influences the quality of reconstructed images, resulting in a better MI robustness.  

\begin{figure*}[h!]
  \centering
  \begin{adjustbox}{width=1.0\textwidth,center}
  \includegraphics[width=0.95\textwidth]{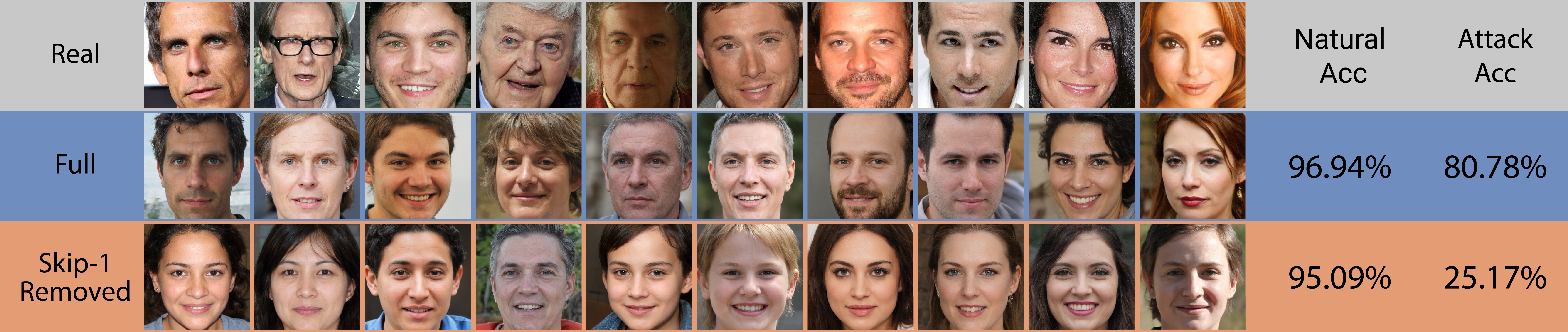}
  \end{adjustbox}

  \label{fig:Viz}
  \vspace{-0.2cm}
  \caption{Qualitative results. Here $T$=MaxViT, $\mathcal{D}_{priv}$=FaceScrub, $\mathcal{D}_{pub}$=FFHQ. The selection of identities and images is entirely random, without any cherry-picking, aiming to provide an unbiased comparison. The results show that the skip connections (i.e., Full) reinforce the MI attack, resulting in the better attack accuracy and reconstructed images that exhibit more visual characteristics of the target identities.}
  \label{fig:Viz}
  \vspace{-0.2cm}
\end{figure*}
\section{Detailed RepVGG Study} \label{Sec:Addtional RepVGG}

In this section, we provide detailed analysis and addtional results for the study of RepVGG \cite{ding2021repvgg} under MI attacks.

\subsection{Removing skip connection in inference-time architecture via RepVGG could not help} \label{Sec:RepVGG}

As discussed in the main manuscript, we seek to explore: \textit{Can RepVGG inference-time architecture (without skip connections) mitigate vulnerability to MI attacks?}

We denote the training-time multi-branch architecture by $T_{RepVGG}$, and the inference-time plain architecture by
$\widehat{T_{RepVGG}}$. In what follows, we provide analytical and empirical analysis to show that \textbf{the gradients in ${\bm{\widehat{T_{RepVGG}}}}$ under MI attacks and that in $\bm{T_{RepVGG}}$ are the same. Therefore, removing skip connection in the inference-time architecture $\bm{\widehat{T_{RepVGG}}}$ could not help mitigate vulnerability to MI.}

Specifically, in the training-time multi-branch architecture $T_{RepVGG}$, there are several blocks. Each block includes three branches: a $3 \times 3$ conv kernel, a $1 \times 1$ conv kernel, and a skip connection. The skip connection helps mitigate gradient vanishing. We denote the $i^{th}$ $T_{RepVGG}$ block  as $z_{i+1} = f_i(z_i)$, which is shown in Eq.~\ref{eq:RepVGG_training}, where $z_i$ represents the input of the $i^{th}$ block, $W^{(k)}$ represents the weight of the $k \times k$ conv kernel (where $k=0$ indicates additive skip connections), and $\mu^{(k)}, \sigma^{(k)}, \gamma^{(k)}, \beta^{(k)}$ represent the accumulated mean, standard deviation, learned scaling factor, and bias of the Batch Normalization (BN) \cite{ioffe2015batch} layer following the $k \times k$ conv kernel. The $i^{th}$ $T_{RepVGG}$ block is \cite{ding2021repvgg}: 

\begin{equation}
\begin{aligned}
    z_{i+1} = f_{i}(z_i) &= BN(z_i * W^{(3)}, \mu^{(3)}, \sigma^{(3)}, \gamma^{(3)}, \beta^{(3)}) \\
            &\hspace{4mm} + BN(z_i * W^{(1)}, \mu^{(1)}, \sigma^{(1)}, \gamma^{(1)}, \beta^{(1)}) \\
            &\hspace{4mm} + BN(z_i, \mu^{(0)}, \sigma^{(0)}, \gamma^{(0)}, \beta^{(0)}) 
\end{aligned}
\label{eq:RepVGG_training}
\end{equation}

After the training phase, RepVGG uses reparameterization to convert a $T_{RepVGG}$ block into a $\widehat{T_{RepVGG}}$ block $\widehat{z_{i+1}} = \widehat{f_i} ({z_i})$, which is a plain conv layer \cite{ding2021repvgg}:

\begin{equation}
\begin{aligned}
    \widehat{z_{i+1}} = \widehat{f_i} ({z_i})&= {z_i} * \widehat{W} + \widehat{b}  \\
\end{aligned}
\label{eq:RepVGG_inference0}
\end{equation}

We  show that \textbf{the outputs of a $\bm{T_{RepVGG}}$ block and a $\bm{\widehat{T_{RepVGG}}}$ block are equal despite their differences in architectures}, i.e., ${f_i}({z_i}) = \widehat{f_i}({z_i})$. Therefore, the gradients in a $T_{RepVGG}$ block, ${\partial f_i}/{\partial z_i}$, are the same as that in a $\widehat{T_{RepVGG}}$ block, ${\partial \widehat{f_i}}/{\partial {z_i}}$.

To show ${f_i}({z_i}) = \widehat{f_i}({z_i})$, we note that $\widehat{W}$ and $\widehat{b}$ in Eq.~\ref{eq:RepVGG_inference0} are obtained in \cite{ding2021repvgg} with the following reparameterization procedure: After training, the conv and BN in each branch of ${f_i}({z_i})$, with kernel $W^{(k)}$ and BN parameters $\{\mu^{(k)}, \sigma^{(k)}, \gamma^{(k)}, \beta^{(k)}\}$ resp., are replaced by another conv layer with parameters $\{\widehat{W^{(k)}}, \widehat{b^{(k)}}\}$, where \cite{ding2021repvgg}:

\begin{equation}
\begin{aligned}
    \widehat{W^{(k)}} = \frac{\gamma^{(k)}}{\sigma^{(k)}}W^{(k)} \\
    \widehat{b^{(k)}} = \beta^{(k)} - \frac{\mu^{(k)} \gamma^{(k)}}{\sigma^{(k)}}    
\end{aligned}
\label{eq:RepVGG_reparam}
\end{equation}

We remark that $\frac{\gamma^{(k)}}{\sigma^{(k)}}W^{(k)}$ in Eq. \ref{eq:RepVGG_reparam} is channel-wise multiplication, as $\sigma^{(k)}, \gamma^{(k)}$ are BN parameters, see \cite{ding2021repvgg}. The same is applied to the skip connection branch, as an identity can be viewed as a $1 \times 1$ conv with an identity matrix as the kernel. In \cite{ding2021repvgg}, the two $1 \times 1$ kernels are then zero-padded to $3 \times 3$ kernels. Then, the three kernels are summed together to obtain $\widehat{W}$ and $\widehat{b}$ in Eq. \ref{eq:RepVGG_inference0} (See \cite{ding2021repvgg}). Therefore, Eq. \ref{eq:RepVGG_inference0} can be re-written as:
{\small
\begin{equation}
\begin{aligned}
    \widehat{f_i} (z_i)&= z_i * \widehat{W} + \widehat{b} \\
            &\overset{(a)}{=} z_i * \widehat{W^{(3)}} + \widehat{b^{(3)}} 
             + z_i * \widehat{W^{(1)}} + \widehat{b^{(1)}} 
             + z_i * \widehat{W^{(0)}} + \widehat{b^{(0)}} \\
        &\overset{(b)}{=} z_i * \frac{\gamma^{(3)}}{\sigma^{(3)}} W^{(3)} +
        \beta^{(3)} - \frac{\mu^{(3)} \gamma^{(3)}}{\sigma^{(3)}}   \\
        &\hspace{4mm} + z_i * \frac{\gamma^{(1)}}{\sigma^{(1)}} W^{(1)} +
        \beta^{(1)} - \frac{\mu^{(1)} \gamma^{(1)}}{\sigma^{(1)}}  \\
        &\hspace{4mm} + z_i * \frac{\gamma^{(0)}}{\sigma^{(0)}} W^{(0)} +
        \beta^{(0)} - \frac{\mu^{(0)} \gamma^{(0)}}{\sigma^{(0)}}  \\ 
        &\overset{(c)}{=} (z_i * W^{(3)} - \mu^{(3)}) \frac{\gamma^{(3)}}{\sigma^{(3)}} + \beta^{(3)} \\
        &\hspace{4mm} + (z_i * W^{(1)} - \mu^{(1)}) \frac{\gamma^{(1)}}{\sigma^{(1)}} + \beta^{(1)} \\
        &\hspace{4mm} + (z_i - \mu^{(0)}) \frac{\gamma^{(0)}}{\sigma^{(0)}} + \beta^{(0)} \\
        &\overset{(d)}{=} f_i(z_i)
\end{aligned}
\label{eq:RepVGG_inference1}
\end{equation}}

In $(a)$, we rewrite $\widehat{W}$ as the sum of three kernels, and remove the zero-padded coefficients to obtain the two $1 \times 1$ kernels: $\widehat{W^{(1)}}$ and $\widehat{W^{(0)}}$. In $(b)$, we use Eq. \ref{eq:RepVGG_reparam}. In $(c)$, we re-arrange the terms. In $(d)$, we use the definition of BN:$ BN(z_i, \mu^{(k)}, \sigma^{(k)}, \gamma^{(k)}, \beta^{(k)}) = (z_i - \mu^{(k)}) \frac{\gamma^{(k)}}{\sigma^{(k)}} + \beta^{(k)} $ and follow Eq. \ref{eq:RepVGG_training}. Overall, Eq.~\ref{eq:RepVGG_inference1} shows that $\widehat{f_i}(z_i)=f_i(z_i)$. As a result, the gradients in a $\widehat{T_{RepVGG}}$ block, ${\partial \widehat{f_i}}/{\partial {z_i}}$ are the same as that in a $T_{RepVGG}$ block, ${\partial f_i}/{\partial z_i}$. Consequently, during MI attacks, the gradients in inference-time architecture, $\widehat{T_{RepVGG}}$, are the same as that in training-time architecture $T_{RepVGG}$.

\subsection{Additional results on other RepVGG architectures}

\begin{table}[t]
\caption{\textbf{Additional experimental results of other RepVGG architectures} \cite{ding2021repvgg}. We strictly follow PPA \cite{struppek2022plug} for the attack setup and evaluation. Here $\mathcal{D}_{priv}$ = Facescrub, $\mathcal{D}_{pub}$ = FFHQ. We report the natural accuracy (Acc), attack accuracy (AttAcc) given in \% and the distance between the reconstructed features and private training features computed using Evaluation Model $\delta_{eval}$ and FaceNet Model \cite{schroff2015facenet} $\delta_{face}$. Across all RepVGG architectures, we find that despite the removal of skip connections, RepVGG inference-time architecture  remains  as vulnerable to MI attacks as the training-time architecture}
\label{tab:RepVGG-Other}
\setlength{\tabcolsep}{0.9em}
  \begin{adjustbox}{width=0.8\columnwidth,center}
  \centering
    \begin{tabular}{cccccc}
    \hline
                               & \textbf{Architecture} \T\B   & \textbf{Acc}   & \textbf{AttAcc} & \boldmath{$\delta_{eval}$}   & \boldmath{$\delta_{face}$} \\ \hline
    \multirow{2}{*}{RepVGG-A1} & Training-time \T\B  & 95.34 & 87.85 & 122.37 &	0.7600    \\
                               & Inference-time \T\B & 95.34 & 88.04 &	122.36 & 0.7614   \\ \hline
    \multirow{2}{*}{RepVGG-A2} & Training-time \T\B  & 95.25 &  87.50	& 121.90 &	0.7697   \\
                               & Inference-time \T\B & 95.25 & 87.29 &	121.75 &	0.7693     \\ \hline
    \multirow{2}{*}{RepVGG-B0} & Training-time \T\B  & 95.50 & 90.24 &	120.32 & 0.7455 	  \\
                               & Inference-time \T\B & 95.50 & 90.09	& 120.17 &	0.7445   \\ \hline
    \multirow{2}{*}{RepVGG-B1} & Training-time \T\B  & 95.65 &  84.29 &	124.98 &	0.7650  \\
                               & Inference-time \T\B & 95.65 & 84.76 &	124.77 &	0.7652   \\ \hline
    \end{tabular}      
  \end{adjustbox}
\end{table}

In addition to the empirical results on RepVGG-A0/B3/D2 in the main manuscript, we provide additional empirical results on RepVGG-A1/A2/B0/B1 in Tab.~\ref{tab:RepVGG-Other}. Overall, the results are consistent with those in the main manuscript, where the inference-time architecture with skip connections removed via RepVGG remains as vulnerable to MI attacks as the training-time architecture.

\section{Additional MI-resilient Architectures} \label{Sec:Additional_Ours}

\subsection{Skip Connection Scaling Factor (SSF) for concatenative skip connection for DenseNets}

\begin{figure*}[ht]
  \centering
  \begin{adjustbox}{width=1.0\textwidth,center}
  \includegraphics[width=1.0\textwidth]{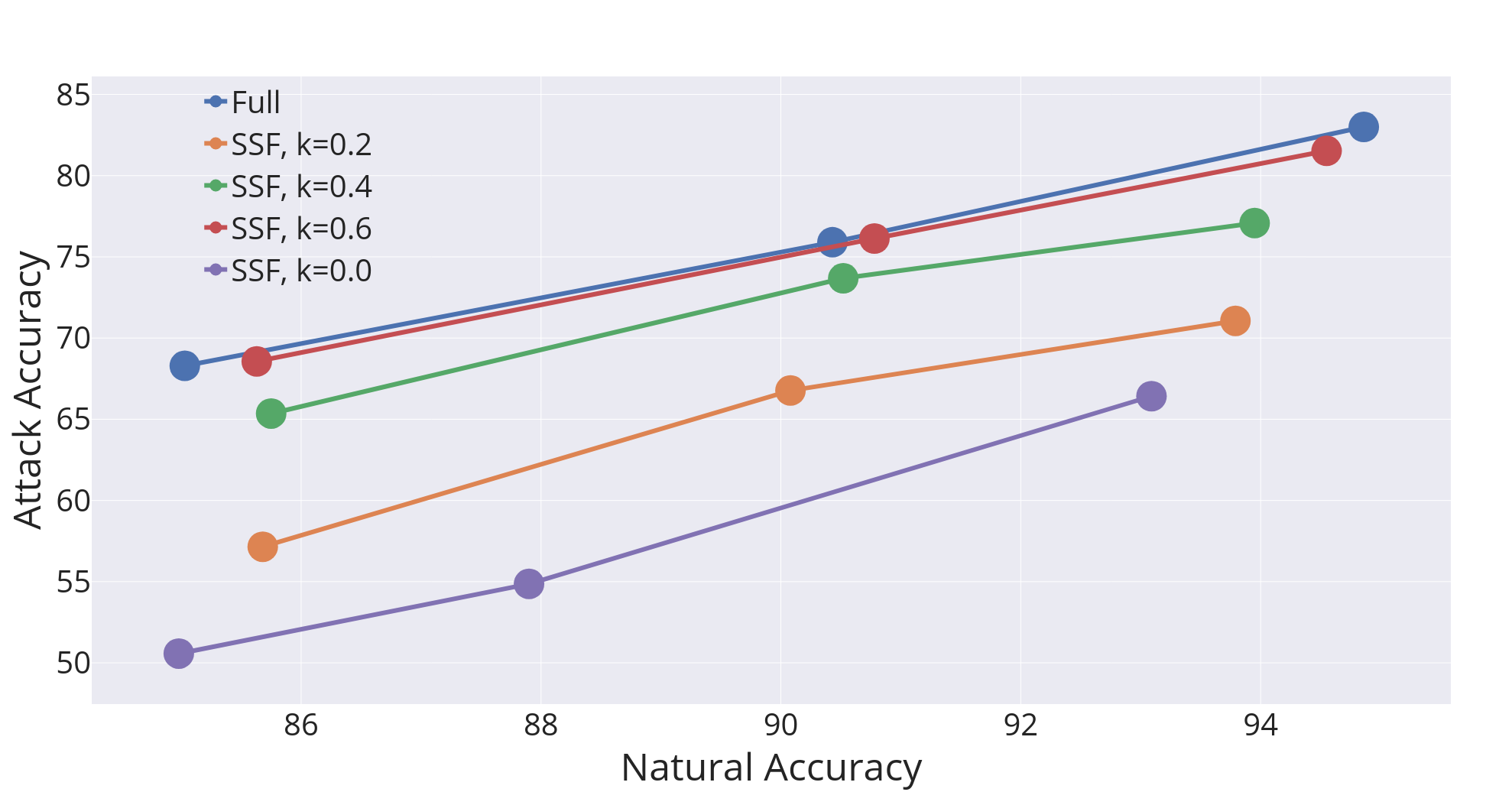}
  \end{adjustbox}
  \caption{
  \textbf{Our ablation study on the effect of $k$ on SSF}. We follow PPA MI setup, where $T$=ResNet-101, $\mathcal{D}_{priv}$=FaceScrub, $\mathcal{D}_{pub}$=FFHQ.
}
  \label{fig:SSF_k}
\end{figure*}

We further propose SSF on top of RoLSS to to improve natural accuracy of the model while maintaining competitive MI robustness. In the main manuscript, we discuss the SSF for additive skip connections as in ResNets, yet SSF is equally applicable to concatenative skip connections as in DenseNets. Concatenative skip connection architectures contain DenseBlocks where input features are concatenated with the output features, before being fed into the next DenseBlock. The scale factor $0 \le k \le 1$ adjusts the signal on the skip connection of the last stage as shown below:

\begin{equation}
    z_{i+1}=[z_i^{scale},g_i(z_i)]
\end{equation}

Here $z_i^{scale}$ is a subset $z_i$ including $k \cdot n$ features from $z_i$, where, $n$ is total number of features of $z_i$. 

Similar to our discussion on SSF for additive skip connections, our SSF generalizes the skip connections, where $k=1$ corresponds to the original skip connection, while $k=0$ is out skip connection removal study. With $k<1$, gradients can be limited during MI attack, and this could degrade MI.

\subsection{MI-resilient architectures are complementary to existing MI defense}

We are the first to explore MI defense from architectural perspective. Therefore, our MI-resilient architectures are complementary to existing MI defense. In this section, we combine the SOTA MI defense, BiDO, with our MI-resilient architectures to further improve MI robustness.

\noindent \textbf{MI setup.} We follow PPA \cite{struppek2022plug} for the MI setup on Facescrub private dataset.

\noindent \textbf{Implementation.} When combining our MI-resilient architecture and BiDO, we strictly follow BiDO. The only difference is that we conduct BiDO on top of our RoLSS architectures.

\begin{wraptable}{r}{6cm}
\vspace{-1.0cm}
\caption{{\bf MI-resilient architectures are complementary to existing MI defense}. $\Delta$ represents the ratio of attack accuracy drop to natural accuracy drop.}
\label{tab:BiDO+Ours}
\setlength{\tabcolsep}{0.7em}
  \begin{adjustbox}{width=0.45\columnwidth,center}
  \centering
        \begin{tabular}{cccc}
        \hline
        \textbf{Defense}    & \textbf{Acc} $\Uparrow$   & \textbf{AttAcc} $\Downarrow$ & $\Delta$ $\Uparrow$ \\ \hline
        No Def.    & 94.86 & 83.00  & -        \\
        BiDO       & 90.31 & 67.07  & 3.50     \\
        BiDO+RoLSS & 89.13 & 41.44  & \textbf{7.25}     \\ \hline
        \end{tabular}  
  \end{adjustbox}
\vspace{-0.7cm}
\end{wraptable}

\noindent \textbf{Experimental result}. The results in Tab.~\ref{tab:BiDO+Ours} show that the trade-off between utility and robustness is much improved with the incorporation of our RoLSS architecture into BiDO. Particularly, the reduction in MI attack accuracy by 25.63\% compared to BiDO alone hile only experiencing a marginal 1\% decrease in natural accuracy. As  pioneering exploration of MI robustness from an architectural perspective, our MI-resilient architecture is complementary to existing regularization-based SOTA MI denfense, such as BiDO.

\subsection{An ablation study on Skip Connection Scaling Factor (SSF)}

In SSF, a scale factor $0 \leq k \leq 1$ is employed to adjust the signal of the skip connection. We conduct an ablation study to examine the impact of $k$ on SSF. We follow the setup of ResNet-101 under PPA attack in the main manuscript with varying $k$. The results in Fig.~\ref{fig:SSF_k} show that as $k$ increase, natural accuracy is more effectively restored. However, larger values of $k$ also lead to a stronger reinforcement of MI attacks.

\subsection{MI-resilient architectures offer flexible control}
\textbf{Our method can have flexible control and our defense performance can be easily improved even further.} Specifically, our proposed RoLSS 
\begin{wraptable}{r}{0.55\columnwidth}
\vspace{-0.8cm}
\caption{(a) RoLSS+ builds on RoLSS by further removing 10\% of skip connections from the second last stage. We show that RoLSS+ can degrade MI attack accuracy more aggressively, which demonstrates that our method offers flexibility and control over privacy utility tradeoff. (b) Our comparison with SOTA MI defenses including MID, DP, and BiDO. The results show that our methods achieve the best MI robustness tradeoff compared to existing MI defenses.}
\label{tab:baseline}
\begin{adjustbox}{width=0.55\columnwidth,center}
  \centering
    \begin{tabular}{ccccc}
    \hline
    \textbf{Architecture} & \textbf{Defense} & \textbf{Acc$\Uparrow$ } & \textbf{AttAcc$\Downarrow$} & \textbf{$\Delta\Uparrow$ } \\ \midrule
    \multirow{8}{*}{ResNet-34}  & No Def.                & 94.69        & 90.78           & -              \\
                                & MID \cite{wang2021improving}         & 91.12        & 46.25           & 12.47          \\
                                & DP  \cite{abadi2016deep}             & 89.66        & 72.19           & 3.70           \\
                                & BiDO \cite{peng2022bilateral}        & 91.66        & 81.98           & 2.90           \\
                                & RoLSS (Ours)           & 91.38        & 71.86           & 5.72           \\
                                & RoLSS+  (Ours)         & 93.49        & 65.78           & 20.83 \\ 
                                & SSF  (Ours)            & 94.21        & 79.79           & 22.90          \\
                                & TTS  (Ours)            & 94.40        & 81.65           & \textbf{31.48} \\ \hline
    \multirow{8}{*}{ResNet-50}  & No Def.                & 94.58        & 82.76           & -              \\
                                & MID  \cite{wang2021improving}        & 89.62        & 66.82           & 3.21           \\
                                & DP   \cite{abadi2016deep}            & 89.97        & 68.89           & 3.01           \\
                                & BiDO \cite{peng2022bilateral}        & 91.12        & 58.41           & 7.04           \\
                                & RoLSS (Ours)           & 92.89        & 68.44           & 8.47  \\
                                & SSF  (Ours)            & 93.05        & 74.79           & 5.21           \\
                                & RoLSS+ (Ours)          & 92.51        & 64.50           & \textbf{8.82}  \\ 
                                & TTS  (Ours)            & 93.56        & 77.21           & 5.44           \\ \hline
    \multirow{8}{*}{ResNet-101} & No Def.                & 94.86        & 83.00           & -              \\
                                & MID  \cite{wang2021improving}        & 90.85        & 52.61           & 7.58           \\
                                & DP  \cite{abadi2016deep}             & 91.36        & 74.88           & 2.32           \\
                                & BiDO \cite{peng2022bilateral}        & 90.31        & 67.07           & 3.50           \\
                                & RoLSS (Ours)           & 92.40        & 58.68           & 9.89           \\
                                & RoLSS+ (Ours)          & 91.05        & 52.74           & 7.94           \\
                                & SSF  (Ours)            & 93.79        & 71.06           & \textbf{11.16} \\
                                & TTS  (Ours)            & 94.16        & 77.26           & 8.20           \\ \hline
    \multirow{8}{*}{ResNet-152} & No Def.                & 95.43        & 86.51           & -              \\
                                & MID \cite{wang2021improving}         & 91.56        & 66.18           & 5.25           \\
                                & DP  \cite{abadi2016deep}             & 91.61        & 75.33           & 2.93           \\
                                & BiDO \cite{peng2022bilateral}        & 91.80        & 58.14           & 7.82           \\
                                & RoLSS (Ours)           & 93.00        & 64.98           & 8.86           \\
                                & RoLSS+ (Ours)          & 92.19        & 54.79           & \textbf{9.79}  \\
                                & SSF (Ours)             & 93.79        & 70.71           & 9.63  \\
                                & TTS (Ours)             & 93.97        & 73.59           & 8.85           \\ \hline
    \end{tabular}
\end{adjustbox}
\vspace{-3.0cm}
\end{wraptable}
focuses on removing skip connections in the last stage only, which is the most critical to MI attacks based on our discovery. This can be easily extended to the other stages to offer greater flexibility and control over privacy utility tradeoff. In Tab.~\ref{tab:baseline}, our results show that RoLSS+ can achieve better privacy utility trade-off than RoLSS.

\subsection{Additional Comparison Against Other MI Defenses}
We provide additional baseline comparison with MID and DP, in Tab.~\ref{tab:baseline}. As shown in the main manuscript and Tab.~\ref{tab:baseline}, \textbf{our proposed method achieves the best tradeoff compared to previous SOTA MI defense.} 
\section{User Study} \label{Sec:User Study}

We utilize Amazon MTurk\footnote{https://www.mturk.com} for our user study, where participants are presented with an image of the target class and tasked with choosing the inverted image that closely resembles the target. Survey questions are randomized, and each image pair is displayed for 60 seconds. Our study covers all 530 identities in the FaceScrub Dataset, with each pair assigned to 10 unique individuals. In this user study, images are generated through the PPA attack under the MaxViT configuration (see Sec.~\ref{Sec:ViT} in this Supp.). Each image pair comprises one MI reconstructed from full skip connection architecture and the other from skip connection removed architecture. 
\begin{wraptable}{r}{6cm}
\vspace{-1.0cm}
\caption{\textbf{User study results.} using PPA attack on MaxViT architectures (Full and Skip Connection Removed). The user study results are consistent with Attack Accuracy, which shows that the skip connections reinforce the MI attack.
}
\label{tab:User_Study}
\begin{adjustbox}{width=0.5\columnwidth,center}
\begin{tabular}{cccc}
\hline
 \textbf{Architecture}   \T \B           & \textbf{Acc} $\uparrow$          & \textbf{AttAcc} $\downarrow$   & \textbf{User Preference} $\downarrow$        \\ \hline
 No Def.      \T              & 96.94\%                     & 80.78\%                     & 69.51\%                   \\
                      Ours    \B      & 95.09\%                     & 25.17\%                     & 30.49\%                   \\ \hline
\end{tabular}
\end{adjustbox}
\vspace{-0.5cm}
\end{wraptable}
Results indicate that when skip connections are removed, the reconstructed images tends to be less similar to the target class, with 69.51\% of users identifying images inverted by the full skip connection architecture as more similar to the target. This reinforces our hypothesis that architectures with fewer skip connections consistently reduce MI attack accuracy.

\begin{figure*}[ht]
  \centering
  \begin{adjustbox}{width=1\textwidth,center}
  \includegraphics[width=0.95\textwidth]{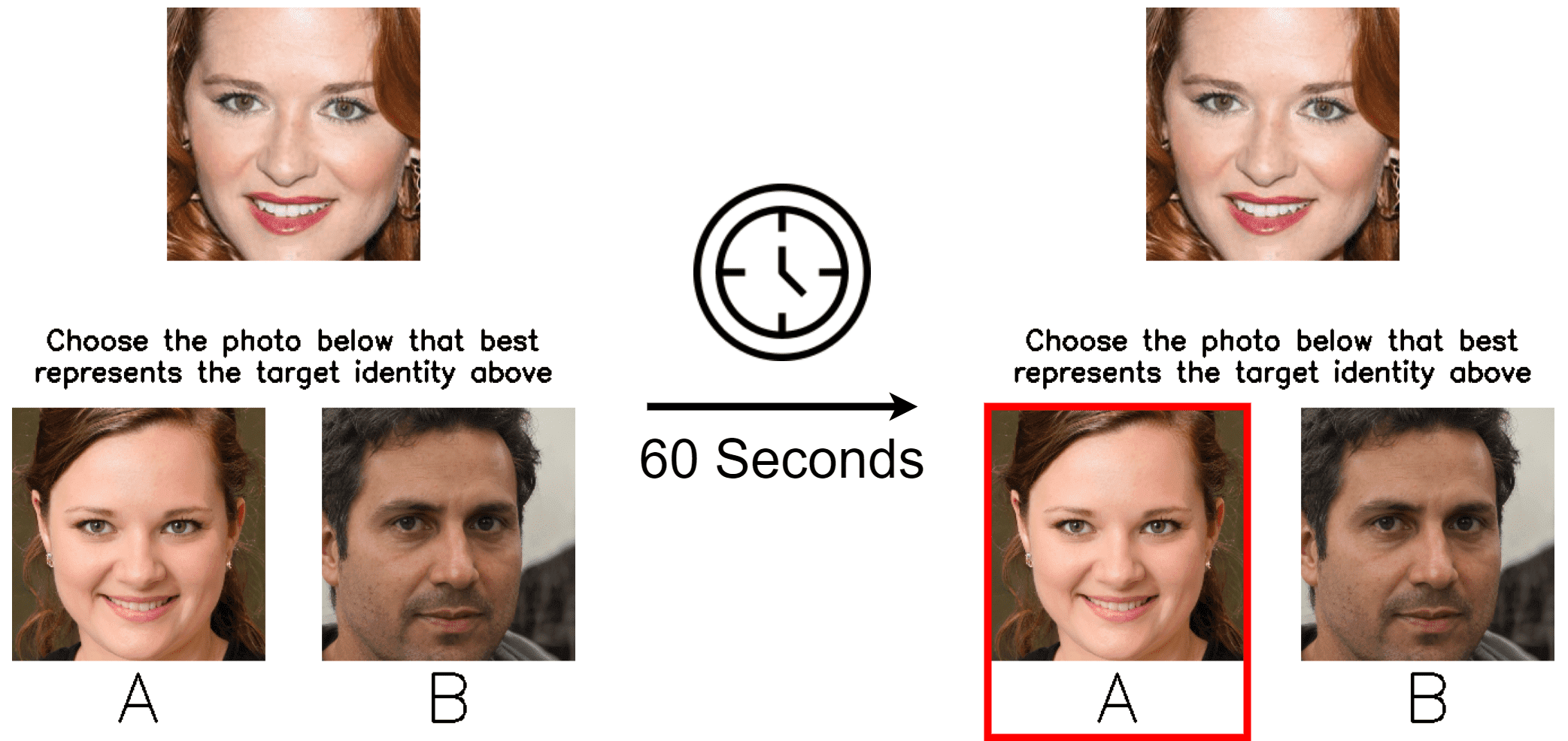}
  \end{adjustbox}
  \caption{
  \textbf{Example of user study survey interface. } The users are asked to choose one image between two options (Full and Skip Connection Removed) that best represents the target identity. For each assignment, users are given 60 seconds to complete the task.
}
  \label{fig:RepVGG_overview}
\end{figure*}

\section{Discussion on architectures without skip connections} \label{Sec:VGG}

In this section, we discuss the model inversion to the architectures without skip connections. For a fair comparison with our study, we conduct high-resolution MI attack experiments on VGG in rebuttal Tab.~\ref{tab:RepVGG-PPA}. We observe that attack accuracy for VGG (49.39\% to 55.57\%) is significantly lower than for architectures with skip connections (82.76\% to 90.78\%, see No. Def results in rebuttal Table \ref{tab:baseline}). This results are consistent with our observation that skip connections reinforce MI attack.

\begin{wraptable}{r}{0.45\columnwidth}
\vspace{-1.25cm}
\caption{Experimental results on high-resolution MI attacks against VGG. We follow PPA for the attack setup and evaluation. Here, $\mathcal{D}_{priv}$=Facescrub and $\mathcal{D}_{pub}$=FFHQ}
\label{tab:RepVGG-PPA}
  \begin{adjustbox}{width=0.45\columnwidth,center}
  \centering
    \begin{tabular}{cccc}
    \hline
    \textbf{Architecture} & \textbf{Acc$\Uparrow$ } & \textbf{AttAcc$\Downarrow$} \\ \midrule
    VGG-16                & 93.70        & 49.39                            \\
    VGG-19                & 93.51        & 55.57                            \\ \hline
    \end{tabular}
    \end{adjustbox}
\vspace{-0.5cm}
\end{wraptable}
\section{Details of Skip Connection Study Setting} \label{training setting}

\begin{figure}[ht]
  \centering
  \begin{adjustbox}{width=1.0\textwidth,center}
  \includegraphics[width=0.95\textwidth]{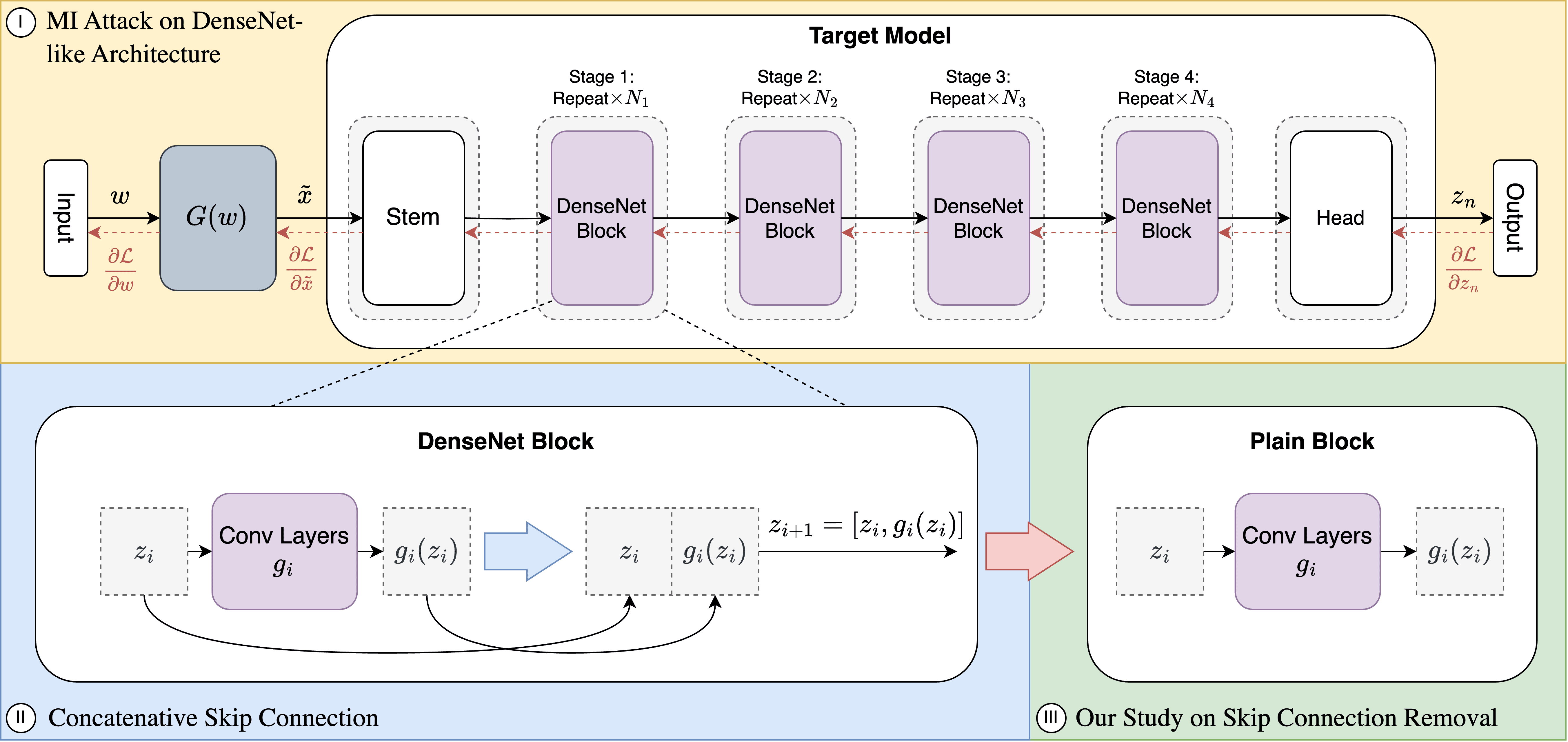}
  \end{adjustbox}
  \caption{
{\bf (I) Illustration of MI attack on DenseNet-like architecture.}
This figure depicts the MI attack framework for SOTA white-box MI attacks \cite{zhang2020secret,chen2021knowledge,wang2021variational,yuan2023pseudo,nguyen_2023_CVPR,struppek2022plug,an2022mirror}, which leverage a generative model $G(.)$ to exploit the target model via gradient descent and backpropagation. Specifically, for each iteration, $  
\tilde{x} = G(w)$ is fed into the target model in the forward pass, and MI loss $\cL$ is computed. In the backward pass, gradients of $\cL$ are computed and back-propagated to obtain $\partial{\cL} / \partial{w}$, which is used to update $w$ to achieve reconstruction of private training data.
 {\bf(II) Concatenative Skip Connection.} During MI attacks, skip connections  enhance backpropagation. We hypothesize that this reinforces MI attacks. In concatenative skip connections,
 input signals 
 are concatenated with the outputs of the current DenseBlock during the feed-forward process.
 {\bf(III) Our study on skip connection removal.} To validate our hypothesis that skip connections could reinforce MI, we study the effect of skip connections on MI by removing  skip connections  within various stages of the target model. We study both  additive skip  connections as discussed  and concatenative skip connections as shown in this sub-figure.   
\textbf{Best viewed in color with zooming in.}
  }
  \label{fig:Architecture_Overview}

\end{figure}

In this section, we provide additional details to the Stage-wise Skip Connection Removal Study as mentioned in the main paper, where we specifically discuss about DenseNet-like architectures that utilizes concatenative skip connections. DenseNet architectures contain DenseBlocks where input features are concatenated with the output features, before being fed into the next DenseBlock as shown in Fig. \ref{fig:Architecture_Overview}-II, where $z_{i+1}=[z_i,g_i(z_i)]$. 


\subsection{Removal of Concatenative Skip Connections} 

To remove concatenative skip connections from DenseNet-like architectures, we remove the concatenation process during the feed forward process within DenseBlocks of these architectures. After removal of these concatenative skip connections, the new latent from subsequent DenseBlocks can be represented as $z_{i+1} = g_i(z_i)$, similar to our study when we remove additive skip connections from ResNet-like architectures. This process is illustrated in Fig.~\ref{fig:Architecture_Overview}-III. 

\subsection{Reproducibility} \label{Sec:Reproducibility}

\noindent \textbf{Hyper-parameters.} We strictly follow the implementations from official source codes \cite{struppek2022plug,chen2021knowledge,nguyen_2023_CVPR}. The details for these hyper-parameter selection are presented in Tab.~\ref{tab:training_T_hyperparam} for training $T$. For a fair comparison, we ensure that the similar training conditions for both architecture with full skip connections and architecture with skip connections removed.

{\renewcommand{\arraystretch}{2.2}
\begin{table}[ht]
    \caption{{\bf Hyper-parameter selection for training $T$.} We follow the hyper-parameter selection from previous works \cite{struppek2022plug,chen2021knowledge,nguyen_2023_CVPR}. We remark that in our skip connection study, the training conditions for architecture with full skip connections and architecture with skip connections removed are similar.}
    \label{tab:training_T_hyperparam}
\centering
    \begin{adjustbox}{width=\columnwidth}
    \begin{tabular}{cccccccc}
    \hline
    \textbf{Architecture}  \T \B        & \textbf{Input Size}               & \textbf{Transformation} & \textbf{Optimizer} & \textbf{LR} & \textbf{LR scheduler} & \textbf{\#Epoch}     & \textbf{Batch Size} \\ \hline
    ResNet-34/50/101/152 \T\B          & \multirow{5}{*}{224 $\times$ 224} & \multirow{2}{*}{RandomResizedCrop}       & \multirow{5}{*}{Adam}               & \multirow{5}{*}{0.001}       & \multirow{5}{*}{MultiStepLR}           & \multirow{5}{*}{100} & \multirow{5}{*}{128}                 \\
    DenseNet-121/161/169/201 \T\B      &                                   & \multirow{2}{*}{ColorJitter}             &                    &             &                       &                      &                     \\
    EfficientNet-B0   \T\B             &                                   & \multirow{2}{*}{RandomHorizontalFlip}    &                    &             &                       &                      &                     \\
    RepVGG-A0/A1/A2/B0/B1/B3/D2 \T\B &                                   &                         &                    &             &                       &                      &                     \\ \hline
    IR152           \T\B               & 64 $\times$ 64                    & RandomHorizontalFlip    & SGD                & 0.01        & -                     &        100              & 64                  \\ \hline
    \end{tabular}
    \end{adjustbox}
\end{table}}

\begin{figure}[h]
    \centering
    \begin{minipage}{0.49\textwidth}
        \centering
        \includegraphics[width=1.1\textwidth]{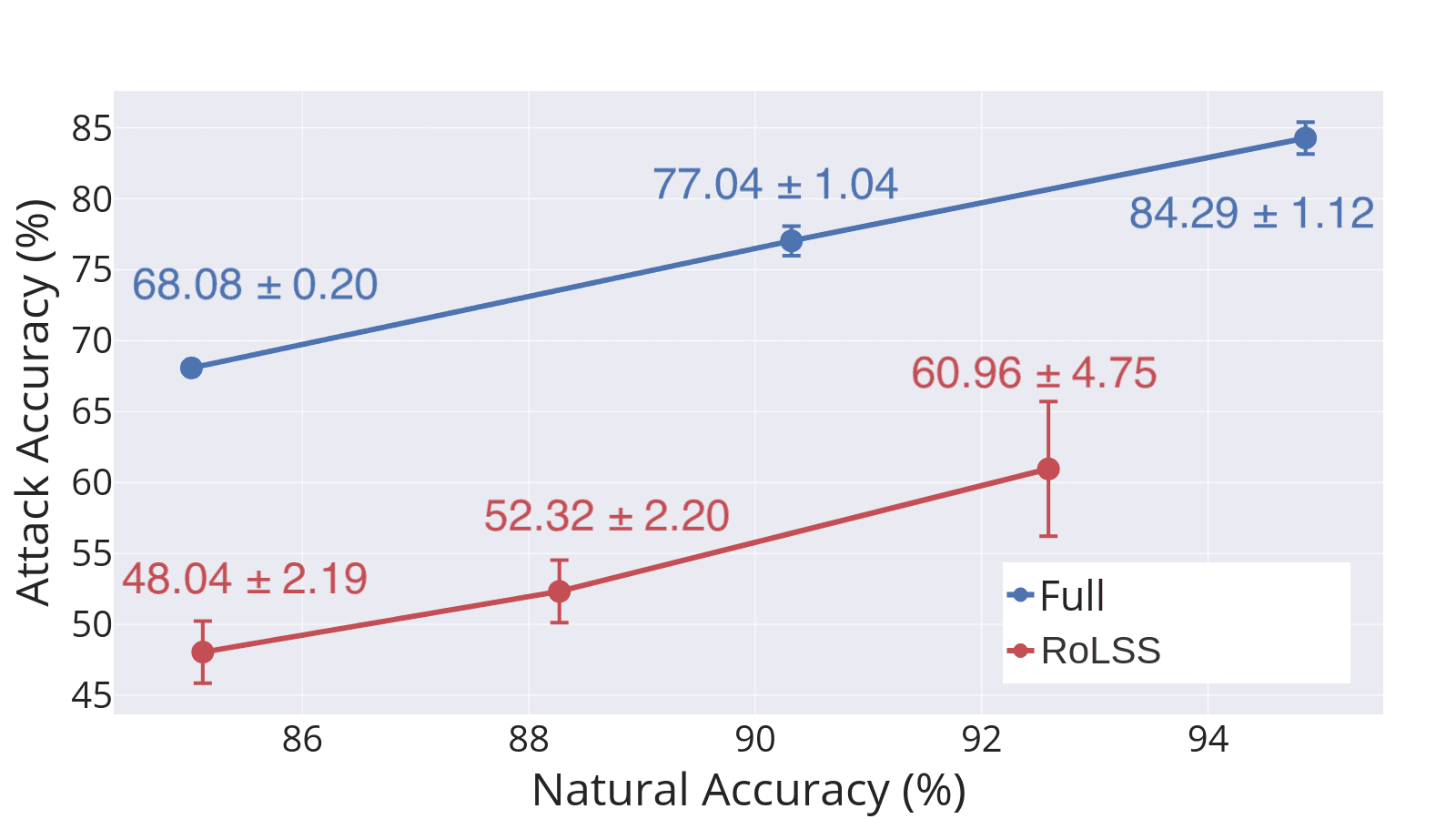}
        \captionof{figure}{Error bar of ResNet-101 Setup}
        \label{Fig:Error_bar_1}
    \end{minipage}
    \hfill
    \begin{minipage}{0.49\textwidth}
        \centering
        \includegraphics[width=1.1\textwidth]{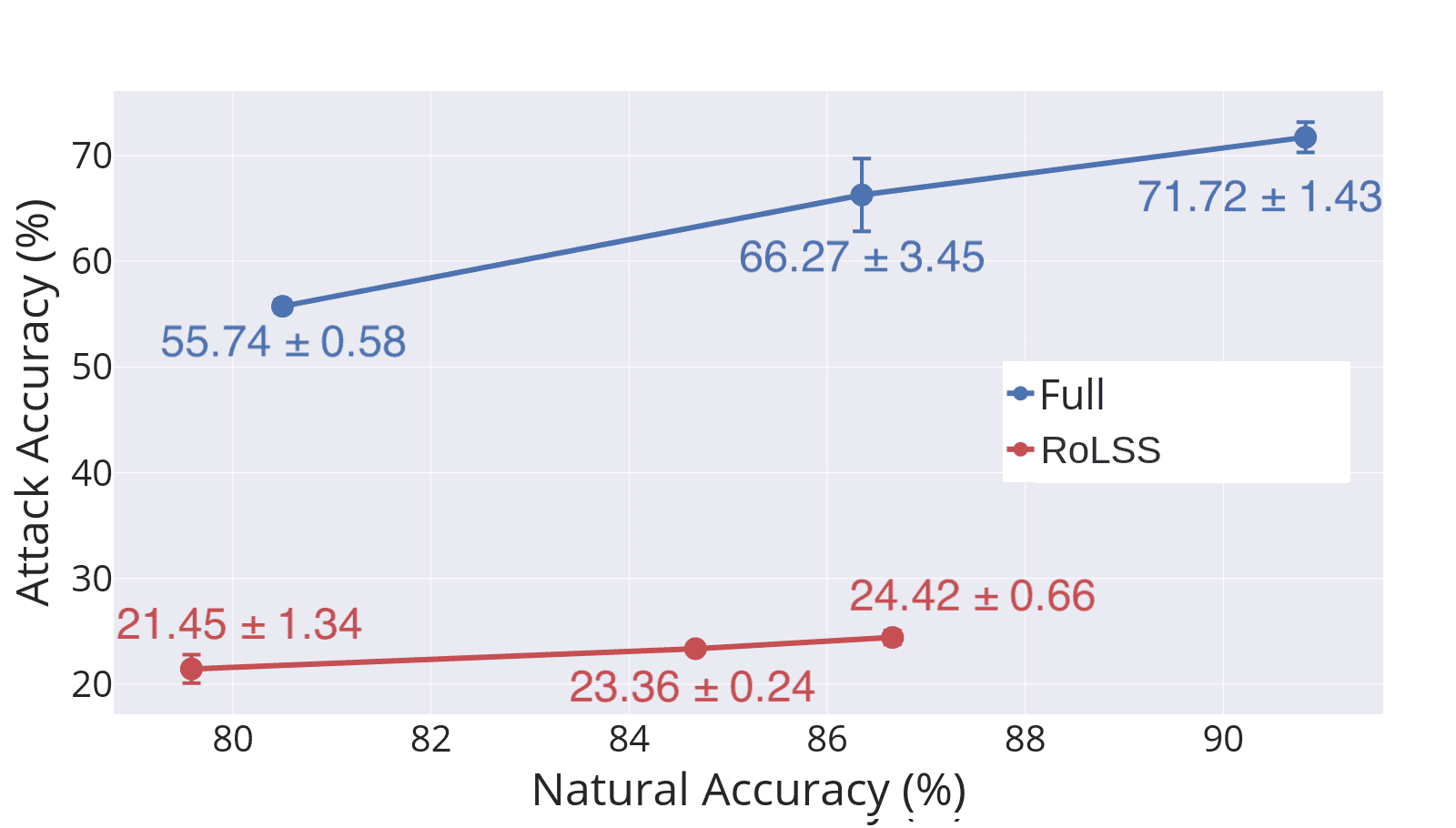}
        \captionof{figure}{Error bar of DenseNet-121 Setup}
        \label{Fig:Error_bar_2}
    \end{minipage}
\end{figure}

\noindent \textbf{Error bar.} To ensure the reproducibility of our findings, we repeat the main experiments reported in the original paper. As MI attacks is very time-consuming, we select key data points from the original paper and evaluate the variations in the results obtained. Specifically, we repeat the stage-wise skip connection removal on ResNet-101 and DenseNet-121 (full and skip-4 removed configurations) for 3 times and report the mean natural accuracy and attack accuracy as well as the standard deviation of the attack accuracy obtained. The results are reported in Fig.~\ref{Fig:Error_bar_1} and Fig.~\ref{Fig:Error_bar_2}. For each experiment, we follow the setup as the previous works.

\section{Details of MI Attack Setup} \label{Sec:MI attack}

\noindent\textbf{Methods for Model Inversion Attacks.} Model Inversion attacks seek to generate synthetic images that capture class-wise characteristics inherent in the private dataset used for training the target classifier. Recent advancements leverage generative adversarial networks (GANs) to enhance attack accuracy, surpassing traditional methodologies. This study meticulously explores generative MI attacks due to their substantial implications for data privacy, with a particular focus on understanding their interaction with skip connections across five different MI attacks.

\textit{KEDMI} \cite{chen2021knowledge} utilizes a MI-specific GAN tailored for MI attacks, integrating knowledge from the target classifier during GAN training. Introducing a new head, the discriminator assumes a dual role by not only discerning between real and fake samples but also predicting the class-wise label of the input. Additionally, the authors advocate for latent distribution modeling to streamline inversion time and enhance the quality of generated samples.

\textit{LOMMA} \cite{nguyen_2023_CVPR} improve prior MI attacks by introducing new logit loss for MI loss and model augmentation concept to avoid MI overfitting.

\textit{PLG-MI} \cite{yuan2023pseudo} employs conditional GANs for MI attacks, effectively segregating the search space for various image classes. Furthermore, the authors incorporate Max-Margin Loss to optimize MI, addressing the vanishing gradient problem inherent in widely used cross-entropy.

\textit{PPA} \cite{struppek2022plug} concentrates on MI attacks tailored for high-resolution images, employing StyleGAN for the inversion task. The proposed SOTA framework highlights its modular nature, allowing for minimal adjustments to the attack setup across diverse architectures and datasets.

\noindent\textbf{Metrics for MI Attack Evaluation.} In alignment with prior research works \cite{chen2021knowledge, nguyen_2023_CVPR, yuan2023pseudo, struppek2022plug}, we utilize Attack Accuracy (AttAcc), K-Nearest-Neighbors Distance (KNN Dist), and distance metrics $\delta_{EvalNet}$ and $\delta_{FaceNet}$ (introduced in PPA) to assess the effectiveness of MI attacks.

\textit{Attack Accuracy (AttAcc):} We utilize a pre-trained evaluation classifier to predict the identities of inverted images, with attack accuracy serving as a metric to gauge the similarity between these inverted images and the target images. To ensure reliability, we employ existing evaluation models from prior studies known for achieving high accuracy scores.

\textit{K-Nearest Neighbors Distance (KNN Dist):} Quantifies the shortest feature distance between the inverted image and the target image, utilizing $l_2$ distance within the penultimate layer of the evaluation model as a measure of feature distance. Consequently, KNN Dist acts as a metric to assess the feature similarity between the reconstructed images and the actual images belonging to the same class.

\textit{$\delta$ Distance:} This metric, introduced in PPA attack, quantifies the similarity between reconstructed images and private training images. It is determined by the $l_2$ distance, measuring the difference in activations between the penultimate layers. Variations of this metric arise based on the model employed to extract these penultimate layer activations. Specifically, $\delta_{EvalNet}$ is calculated using the Evaluation Model, whereas $\delta_{FaceNet}$ is computed utilizing the pre-trained FaceNet \cite{schroff2015facenet}.

\section{Related Work} \label{Sec:Related Work}

{\bf Model Inversion.} The concept of MI was initially studied by Fredrikson et al. \cite{fredrikson2014privacy}, who demonstrated that adversaries could employ machine learning to extract genomic and demographic information about patients from a medical imaging model. This work was later extended to facial recognition in \cite{fredrikson2015model}.  An adversarial model inversion approach was introduced by Yang et al. in \cite{yang2019neural}. This approach utilizes the target classifier as an encoder to generate a prediction vector, which is used as input to a second network for reconstructing the original data.

Since then, several MI studies have been conducted to understand the feasibility and extent of reconstructing private training samples from DNNs.  These studies encompass both MI attacks and MI defense perspectives. 
Firstly, recent works analyzed the limitations of conventional MI objectives and proposed enhancements to MI attacks, where PLGMI \cite{yuan2023pseudo}, LOMMA \cite{nguyen_2023_CVPR}, PPA \cite{struppek2022plug} utilize logit maximization loss, Max-Margin loss \cite{yang2022towards,sriramanan2020guided} or Point Care loss \cite{beardon1999klein}. Other works modified the MI objective to facilitate MI attacks in black-box \cite{han2023reinforcement} and label-only \cite{kahla2022label} scenarios. Secondly, regularization techniques in MI were explored to improve the realism of reconstructed images \cite{zhang2020secret}. Thirdly, advanced MI attacks for high-dimensional data, such as images, examined the effect of distributional priors in guiding MI. Specifically, GMI \cite{zhang2020secret} used a pretrained GAN \cite{goodfellow2014generative} to learn the image structure of an auxiliary dataset with a similar structure to the target image space. Inversion images are then found through the latent vector of the generator. VMI \cite{wang2021variational} offers a probabilistic interpretation of MI, which leads to a variational objective for the attack.  KEDMI \cite{chen2021knowledge} proposed to use a MI specific GAN trained on knowledge from the target model.  PLGMI \cite{yuan2023pseudo} proposed to use conditional GAN \cite{miyato2018spectral} to decouple the search space for different classes of images. For high-resolution MI attacks, MIRROR \cite{an2022mirror} and PPA \cite{struppek2022plug} leverage the power StyleGAN \cite{karras2019style} and perform MI on $\mathcal{W}$ space. Finally, regularizations on the training objective of the target model as methods to defend against MI attacks have been studied in \cite{wang2021improving, peng2022bilateral}. More concretely, MID \cite{wang2021improving} limits the input-output dependency through a mutual information penalization, while BiDO \cite{peng2022bilateral} aims to minimize the dependency (via COCO \cite{gretton2005kernel} or HSIC \cite{gretton2005measuring} measurements) between latent representations and inputs while maximizing the dependency between latent representations and outputs. {\em Despite considerable  progress in MI research, there is a lack of study to understand the effect of DNN architecture design on MI.}

\noindent {\bf Skip connections.} A notorious problem of training very deep networks is that gradients 
could vanish when they reach initial layers of the networks \cite{hanin2018neural,balduzzi2017shattered,glorot2010understanding}. Various efforts have been employed to address this issue, including the utilization of Rectified Linear Units (ReLU) \cite{agarap2018deep}, the implementation of Batch Normalization \cite{ioffe2015batch}, and the application of specialized weight initialization methods \cite{arpit2019initialize}. From the DNNs architecture perspective, adding shortcut connections has been recognized as an effective approach to alleviate the vanishing gradient problem.

The implementation of skip connections in DNNs commonly adopts additive skip connections \cite{he2016deep}, where the output of a previous layer is added to the output of the current layer. This implementation is known for its simplicity and effectiveness. Another prevalent implementation is the concatenative skip connection \cite{huang2017densely}, wherein each layer receives concatenated feature maps from all preceding layers. Through the concatenation operation along the channel dimension, this method ensures a more comprehensive set of features for subsequent layers to process. Noteworthy advanced deep neural networks, such as DenseNet \cite{huang2017densely}, ResNet \cite{he2016deep}, MaxViT \cite{tu2022maxvit}, EfficientNet \cite{tan2019efficientnet, tan2021efficientnetv2}, and others \cite{dosovitskiy2020image, ding2021repvgg, xu2022regnet,liu2022convnet}, leverage skip connections during training to improve their performance.

\section{Limitation} \label{Sec:Limitation}
In our experiments, we employed network architectures commonly used in MI research. Furthermore, we included a very recent network architecture, namely MaxViT. We observed consistent results across the range of network architectures we employed. Meanwhile, the study of additional network architectures may be included.

\section{Ethical Consideration} \label{Sec:Ethical Consideration}
Our study highlights the vulnerability of skip connections to privacy threats. We hope that our findings raise awareness of potential private data leaks associated with high-performance network architectures. We urge further research into privacy-safe network architectures.

%
%

\end{document}